%% file: main.tex
\documentclass[runningheads]{llncs}

\usepackage[mobile]{eccv}

\usepackage{eccvabbrv}

\usepackage{graphicx}
\usepackage{booktabs}

\usepackage{subcaption}

\usepackage[accsupp]{axessibility}
\usepackage{makecell}

\usepackage{hyperref}

\usepackage{orcidlink}
\usepackage{float}

\begin{document}

\makeatletter
\apptocmd{\@maketitle}{

 \centering
 \vspace{-2pt}
 
 \begin{minipage}{0.88\textwidth}
 \centering
\includegraphics[width=\textwidth]{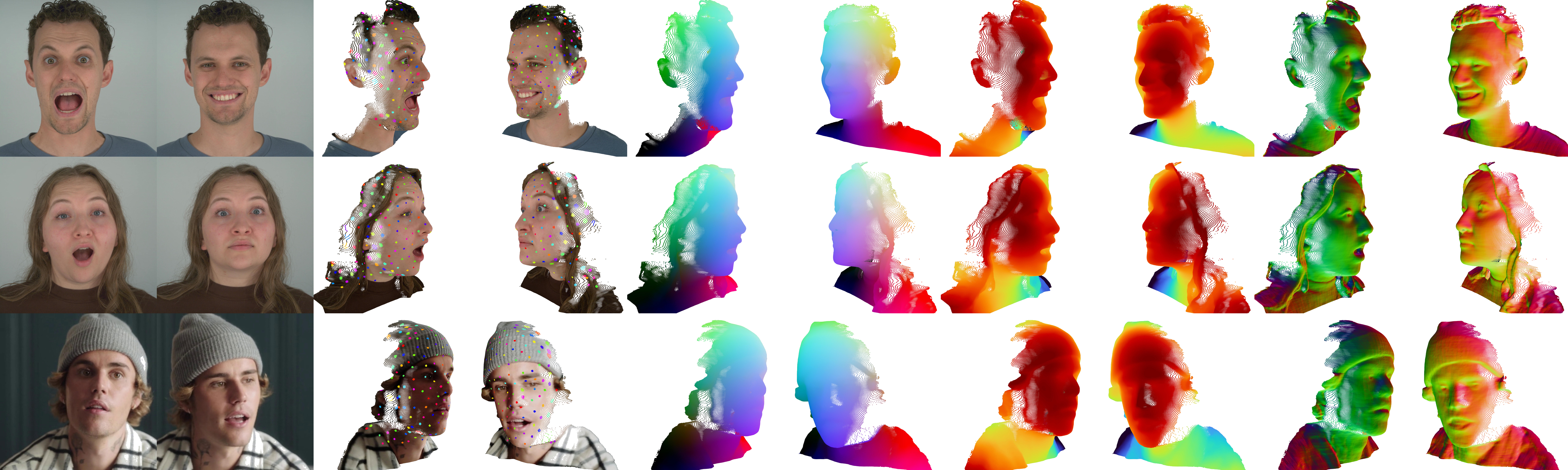}

\captionof{figure}{
\textbf{Face Anything.} Unified 4D facial reconstruction and dense tracking from image sequences via joint prediction of depth and canonical facial coordinates. Left to right: RGB input, 4D reconstruction with tracks, canonical maps, depth maps, and normal maps. Website: \url{https://kocasariumut.github.io/FaceAnything/}
}

  \label{fig:teaser}
  \end{minipage}
  
  \vspace{-18pt}
}{}{}
\makeatother

\title{Face Anything: 4D Face Reconstruction \\ from Any Image Sequence}

\titlerunning{Face Anything}

\author{
Umut Kocasarı\inst{1} \and
Simon Giebenhain\inst{1} \and
Richard Shaw\inst{2} \and
Matthias Nießner\inst{1}
}

\authorrunning{U. Kocasarı et al.}

\institute{
Technical University of Munich, Germany\and
Huawei Noah’s Ark Lab, London, UK
}

\maketitle

\begin{abstract}

Accurate reconstruction and tracking of dynamic human faces from image sequences is challenging because non-rigid deformations, expression changes, and viewpoint variations occur simultaneously, creating significant ambiguity in geometry and correspondence estimation. We present a unified method for high-fidelity 4D facial reconstruction based on canonical facial point prediction, a representation that assigns each pixel a normalized facial coordinate in a shared canonical space. This formulation transforms dense tracking and dynamic reconstruction into a canonical reconstruction problem, enabling temporally consistent geometry and reliable correspondences within a single feed-forward model. By jointly predicting depth and canonical coordinates, our method enables accurate depth estimation, temporally stable reconstruction, dense 3D geometry, and robust facial point tracking within a single architecture. We implement this formulation using a transformer-based model that jointly predicts depth and canonical facial coordinates, trained using multi-view geometry data that non-rigidly warps into the canonical space. Extensive experiments on image and video benchmarks demonstrate state-of-the-art performance across reconstruction and tracking tasks, achieving approximately 3$\times$ lower correspondence error and faster inference than prior dynamic reconstruction methods, while improving depth accuracy by 16\%. These results highlight canonical facial point prediction as an effective foundation for unified feed-forward 4D facial reconstruction.

\end{abstract}

\section{Introduction}
\label{sec:intro}

Reconstructing and tracking dynamic human faces from image sequences is a fundamental problem in computer vision with applications in virtual avatars, telepresence, animation, and human–computer interaction. Systems must recover detailed dynamic geometry and establish consistent correspondences across time from unconstrained image sequences. Despite recent progress in feed-forward 3D reconstruction~\cite{wang2025vggt,wang2025pi,lin2025da3}, achieving temporally consistent 4D facial reconstruction with reliable tracking remains challenging. Faces exhibit complex non-rigid deformations caused by expressions and head motion while preserving fine geometric structures such as wrinkles, hair, and mouth interiors. Existing approaches often struggle to maintain both geometric fidelity and consistent correspondences under large expressions, extreme viewpoints, and long sequences.

Recovering dynamic facial geometry from image observations is fundamentally underconstrained and therefore requires strong learned priors. In addition to reconstructing geometry, 4D facial understanding requires correspondences across time, introducing an additional representational challenge. Most existing methods formulate tracking as predicting how points move across frames~\cite{st4rtrack2025,sucar2025dynamicpm,sucar2025vdpm}, requiring reasoning about mappings between frame pairs. Such motion-based formulations become increasingly difficult to learn as motion complexity and sequence length increase and require large amounts of supervision to obtain stable correspondences. In contrast, representing faces in a normalized canonical space is simpler, since canonical geometry is largely shared across poses and expressions and provides a stable reference for correspondence estimation.

Existing approaches address these challenges using different forms of geometric priors. Learning-based reconstruction models such as DA3~\cite{lin2025da3} learn geometric priors from large-scale data and achieve monocular and multi-view depth prediction, but do not establish correspondences required for tracking. Face-specific predictors such as DAViD~\cite{saleh2025david} and Sapiens~\cite{khirodkar2024sapiens} provide accurate single-image facial geometry estimation but operate independently per frame and therefore lack temporal consistency. Correspondence-based methods such as Pixel3DMM (P3DMM)~\cite{giebenhain2025pixel3dmm} and V-DPM~\cite{sucar2025vdpm} estimate tracking, but parametric representations limit geometric detail while motion-based formulations require multiple forward passes and high computational cost. As a result, existing approaches treat reconstruction and correspondence estimation as separate problems or rely on representations that do not scale well to detailed dynamic faces.

In this work, we introduce a unified method for high-fidelity facial reconstruction and tracking based on canonical map prediction. Instead of predicting frame-to-frame motion, our method predicts a dense canonical map assigning each pixel a canonical facial coordinate in a normalized pose and expression space. Correspondences are obtained through nearest-neighbor search in canonical space, transforming tracking into a canonical reconstruction problem. This representation naturally enforces temporal consistency while remaining efficient to compute and provides normalized geometry suitable for downstream tasks such as animation and avatar generation. We implement the proposed formulation with a transformer-based architecture that jointly predicts depth and canonical facial points. The network uses a DPT-style head~\cite{ranftl2021vision} to process multiple input images simultaneously. Because learning canonical representations requires dense supervision that is largely absent from existing datasets, we construct a dataset based on NeRSemble~\cite{kirschstein2023nersemble} with high-quality multi-view reconstructions and canonical correspondences aligned using FLAME~\cite{li2017flame}. Extensive experiments demonstrate state-of-the-art performance in face depth estimation, temporally stable video depth prediction, dense 4D visible-surface reconstruction, and facial point tracking. Compared with prior approaches such as V-DPM~\cite{sucar2025vdpm}, our method achieves superior reconstruction and tracking accuracy while requiring less computation and memory.

In summary, this paper makes the following contributions:
\begin{itemize}
\item We propose a novel transformer-based method for unified 4D facial reconstruction and tracking. Unique properties of the face domain enable us to exploit canonical position map prediction, in addition to depth and ray maps, allowing for temporally stable correspondences.

\item To supervise our novel formulation, we introduce a large-scale dynamic 3D face dataset, including canonicalized representations, obtained from high-quality MVS reconstructions and FLAME tracking.

\item We achieve state-of-the-art performance on facial single-image and video depth estimation, 4D visible-surface reconstruction, and 3D point tracking.
\end{itemize}

\section{Related Work}
\label{sec:related_work}

\textbf{Static 3D and Feed-Forward Reconstruction.}
Classical 3D reconstruction relies on structure-from-motion and multi-view stereo pipelines that estimate cameras and dense geometry through global optimization~\cite{schoenberger2016sfm, schoenberger2016mvs}, with systems such as COLMAP~\cite{schoenberger2016sfm} remaining standard for static scenes. Neural implicit representations such as NeRF and its extensions~\cite{mildenhall2020nerf, barron2021mipnerf, chen2022tensorf, mueller2022instant, martinbrualla2020nerfw, wang2022sparsenerf, niemeyer2021regnerf}, together with explicit formulations including 3D Gaussian Splatting~\cite{kerbl3Dgaussians, chen2025dashgaussian, mallick2024taming3dgs, meuleman2025onthefly}, enable reconstruction via differentiable rendering but require scene-specific optimization.

Recent work focuses on feed-forward geometric inference from images, including correspondence-based reconstruction~\cite{wang2023dust3r, leroy2024mast3r}, unified multi-view transformer models~\cite{wang2025vggt, wang2025pi, keetha2026mapanything}, feed-forward splatting approaches~\cite{charatan2023pixelsplat, xu2024depthsplat, chen2024mvsplat, jiang2025anysplat, moreau2026offthegrid, ye2024noposplat}, and large-scale geometry foundation models such as Depth Anything~\cite{lihe2024da, lihe2024da2, lin2025da3}. While enabling accurate feed-forward reconstruction, these approaches do not explicitly predict dense correspondences required for tracking deformable objects.

\textbf{Dynamic 4D Reconstruction and Tracking.}
Dynamic extensions to neural implicit representations model non-rigid motion through time-conditioning and deformation fields~\cite{pumarola2020dnerf, park2021nerfies, park2021hypernerf}, while dynamic Gaussian representations enable efficient time-dependent scene modeling and primitive tracking~\cite{luiten2023dynamic, wu20244dgs, li2023spacetime, Shaw2024swings, huang2023sc}. 

Recent work moves toward feed-forward dynamic reconstruction, including dynamic splatting and online reconstruction methods~\cite{xu20254dgt, wang2025movies, wu2026streamsplat} and multi-view geometric frameworks predicting temporally aligned geometry or point maps across frames~\cite{zhang2024monst3r, lu2024align3r, cut3r, wang2025pi, zhou2026page4d, jiang2025geo4d}. Dynamic Point Map representations recover scene motion in a feed-forward manner~\cite{sucar2025dynamicpm, st4rtrack2025, sucar2025vdpm}. Tracking-any-point methods estimate correspondences purely in image space~\cite{bian2023contextpips, doersch2023tapir, karaev23cotracker, li2024taptr, cho2024locotrack}. While dynamic reconstruction methods recover geometry and tracking approaches estimate correspondences, neither explicitly predicts dense canonical coordinates that provide identity-consistent alignment under non-rigid deformation.

\textbf{Monocular Face Reconstruction and Parametric Models.}
Monocular face reconstruction typically relies on parametric 3D Morphable Models (3DMMs), such as the Basel Face Model and FLAME~\cite{paysan2009bfm, gerig2018face, li2017flame}, representing identity and expression in low-dimensional subspaces. Optimization- and learning-based methods fit model parameters directly to images or video~\cite{blanz1999morphable, tewari2017mofa, richardson20173d, deng2019accurate}. Subsequent approaches further improve geometric detail and correspondence estimation~\cite{feng2018prnet, danecek2022emoca, giebenhain2025pixel3dmm, wang2022faceverse, ming2025vggtface} but remain constrained by fixed parametric topology or limited temporal modeling. Recent large-scale approaches such as DAViD~\cite{saleh2025david} and Sapiens~\cite{khirodkar2024sapiens} improve single-image facial geometry prediction but do not enforce temporal consistency or dense correspondences.

Neural head avatar methods IMavatar~\cite{zheng2022imavatar} and Neural Head Avatars~\cite{grassal2022neural} reconstruct subject-specific dynamic heads via optimization. Splatting–based methods~\cite{dhamo2023headgas, xu2023gaussianheadavatar, zhao2024psavatar, wang2024gaussianhead, chen2024monogaussianavatar, zhou2024headstudio, shaw2025ico3d, wu2026uika} represent animatable heads using Gaussians for high-fidelity real-time rendering. However, these approaches rely on parametric representations or subject-specific optimization, leaving geometry and correspondence estimation largely decoupled. In contrast, our method predicts dense canonical coordinates supervised by parametric alignment but represented non-parametrically. This reduces correspondence estimation to reconstruction in canonical space, enabling unified feed-forward 4D reconstruction and tracking.

\section{Method}
\label{sec:methodology}

\subsection{Architecture}
\label{sec:methodology_framework}

\begin{figure*}[t]
\centering
\includegraphics[width=\textwidth]{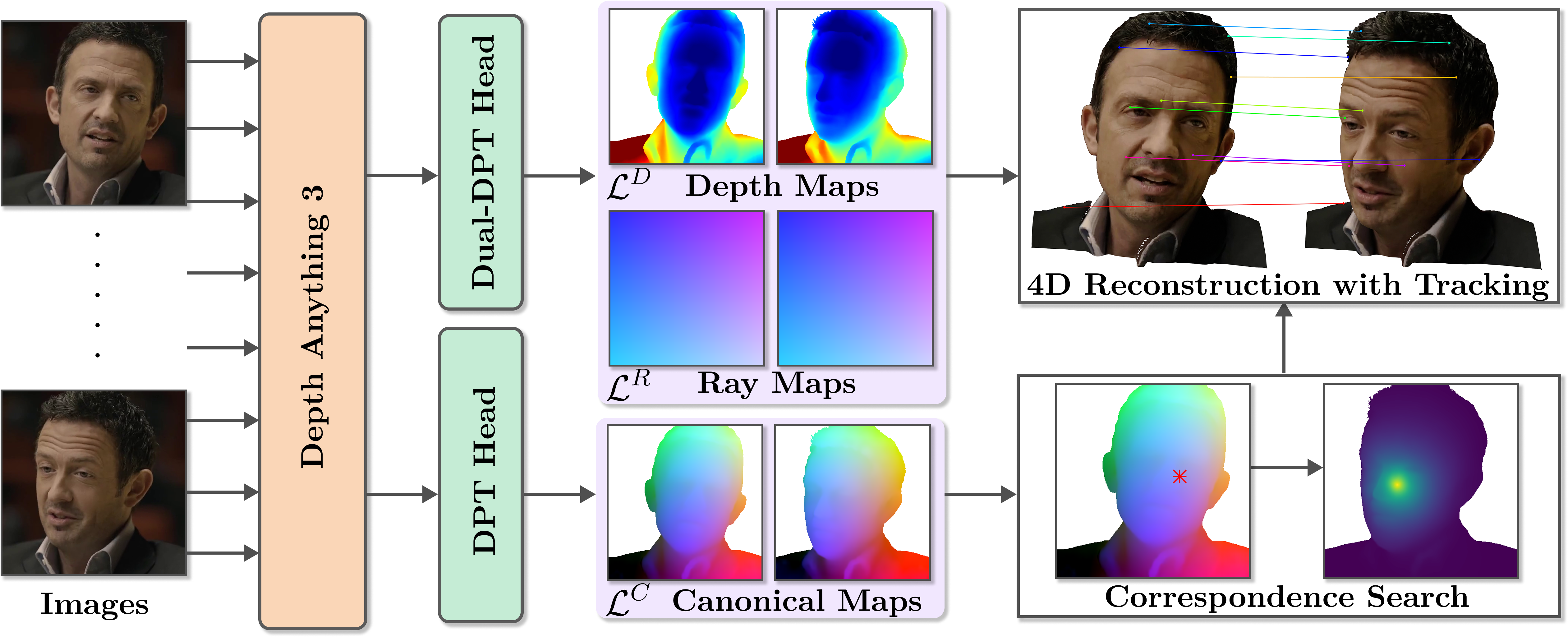}

\caption{
\textbf{Architecture overview.} Given image sequences, our method jointly predicts depth and canonical facial maps to enable dense 4D visible-surface reconstruction and tracking. Dense correspondences are established in canonical space, producing temporally consistent geometry and point trajectories.
}
\label{fig:framework_method}

\end{figure*}

We reconstruct dynamic faces and establish dense correspondences using a single feed-forward network, as illustrated in \cref{fig:framework_method}. Instead of predicting frame-to-frame motion or deformation fields, we formulate correspondence estimation as \textbf{canonical map prediction}, where each pixel is mapped to a canonical facial coordinate. This formulation requires only one forward pass followed by efficient canonical-space matching, making it more efficient than motion-based approaches that require multiple evaluations. It also simplifies learning, since canonical geometry is normalized and structurally similar across poses and expressions, whereas deformation targets vary significantly with viewpoint and motion.

Our method employs a transformer-based architecture that predicts depth, ray maps, and canonical maps from one or more input images. The design is related to DA3~\cite{lin2025da3}, which predicts depth and ray maps. To support canonical prediction, we incorporate a DPT head~\cite{ranftl2021vision}. Given input images $\mathcal{I}=\{I_i\}_{i=1}^{N}$, the network predicts depth maps $D_i\in\mathbb{R}^{H\times W}$, ray maps $R_i\in\mathbb{R}^{H\times W\times3}$, and canonical maps $C_i\in\mathbb{R}^{H\times W\times3}$:

\begin{equation}
(D_i,R_i,C_i)=f_\theta(I_1,\dots,I_N).
\end{equation}

\textbf{Training.}
We train the model in two stages. First, the architecture is pretrained on DAViD~\cite{saleh2025david} using monocular input to learn facial geometric priors, where depth is supervised using a masked $L_1$ regression loss. After pretraining, we add the canonical prediction head and finetune the full network on the dataset described in \cref{sec:methodology_dataset_creation}. The network predicts depth, ray, and canonical maps jointly.

To reduce the domain gap between multi-view capture and monocular video, training alternates between two sampling strategies: (1) multi-view images from a single timestamp and (2) single-camera images across multiple timestamps. This improves generalization to both reconstruction and tracking scenarios.

\textbf{Correspondence Estimation.}
Dense correspondences are obtained by nearest-neighbor search in canonical space. Given images $I_i$ and $I_j$, a pixel $\mathbf{p}$ in $I_i$ corresponds to

\begin{equation}
\mathbf{q} =
\arg\min_{\mathbf{q}' \in \Omega_j}
\left\|
C_i(\mathbf{p}) -
C_j(\mathbf{q}')
\right\|_2,
\end{equation}

where $\mathbf{q}$ denotes the corresponding pixel in image $I_j$ and $\Omega_j$ is the set of pixels in $I_j$.

Nearest-neighbor search is implemented using KD-Tree~\cite{bentley1975multidimensional} implementation and typically requires less than $0.2$ seconds per image pair on a single CPU. The process is fully parallelizable across timestamps and can be accelerated via spatial downsampling with negligible accuracy loss.

Compared to deformation-based approaches such as V-DPM~\cite{sucar2025vdpm}, canonical map prediction is both easier to learn and more efficient, producing correspondences in a single forward pass while maintaining high reconstruction fidelity.

\subsection{Loss Formulation}
\label{sec:losses}

For each prediction type $X\in\{D,R,C\}$ with ground truth $X^*$, we use regression, confidence-weighted regression, and gradient losses:

\begin{eqnarray}
\mathcal{L}_{\text{reg}}^X &=&
\frac{1}{|\Omega|}
\sum_{\mathbf{p}\in\Omega}
|X(\mathbf{p})-X^*(\mathbf{p})|
\\
\mathcal{L}_{\text{conf}}^X &=&
\frac{1}{|\Omega|}
\sum_{\mathbf{p}\in\Omega}
\left(
\gamma |X(\mathbf{p})-X^*(\mathbf{p})|W_X(\mathbf{p})
-\alpha\log W_X(\mathbf{p})
\right)
\\
\mathcal{L}_{\text{grad}}^X &=&
\frac{1}{|\Omega|}
\sum_{\mathbf{p}\in\Omega}
\left(
|\nabla_x E_X(\mathbf{p})|
+
|\nabla_y E_X(\mathbf{p})|
\right),
\quad E_X=X-X^*
\\
\mathcal{L} &=&
\lambda_C
\left(
\mathcal{L}_{\text{reg}}^C +
\mathcal{L}_{\text{conf}}^C +
\mathcal{L}_{\text{grad}}^C
\right)
+
\sum_{X\in\{D,R\}}
\left(
\mathcal{L}_{\text{reg}}^X +
\mathcal{L}_{\text{conf}}^X +
\mathcal{L}_{\text{grad}}^X
\right)
\end{eqnarray}

where $\Omega$ denotes valid pixels, $W_X(\mathbf{p})$ is the predicted confidence, $\alpha=0.2$ and $\gamma=1$ are scalar weighting parameters, and $\lambda_C=5$ weights the canonical map losses. $\mathcal{L}_{\text{reg}}$ drives learning, $\mathcal{L}_{\text{grad}}$ enforces smoothness, and $\mathcal{L}_{\text{conf}}$ down-weights uncertain regions and yields per-pixel confidence; canonical losses receive a higher weight since the canonical head is trained from scratch.

\subsection{Dataset Creation}
\label{sec:methodology_dataset_creation}

Learning canonical facial correspondences requires supervision largely absent from existing datasets. To address this, we construct a dataset based on NeRSemble~\cite{kirschstein2023nersemble}, which provides synchronized multi-view videos with calibrated cameras. We first pretrain on DAViD~\cite{saleh2025david} to learn human-specific priors, then finetune on the NeRSemble-based dataset to learn detailed facial geometry and canonical correspondences. NeRSemble contains multi-view recordings captured with 16 cameras across diverse subjects, poses, and expressions. We use 414 subjects and approximately 20k timestamps across multiple sequences, corresponding to roughly 320k images.

Rather than sampling timestamps uniformly, we select frames that maximize diversity in facial expressions and head poses. To achieve this, we run MediaPipe~\cite{lugaresi2019mediapipe} on the frontal camera view to estimate facial blendshape parameters and head poses for all sequences. For each subject, we select 50 timestamps using farthest point sampling to ensure diverse coverage. We separately sample 40 timestamps based on blendshape parameters and 10 timestamps based on pose parameters in order to balance expression and pose variation.

For each selected timestamp, we reconstruct geometry using COLMAP~\cite{schoenberger2016sfm} from all 16 camera views. Reconstruction is performed on images downsampled by a factor of four for computational efficiency. We retain only reconstructions where all views are successfully registered, as incomplete registrations often indicate unreliable geometry. The resulting reconstructions provide multi-view consistent depth maps and dense point clouds with detailed facial geometry. To improve reconstruction quality in challenging regions such as hair, we adjust COLMAP hyperparameters to obtain more complete point coverage.

To establish canonical correspondences, we perform FLAME-based tracking~\cite{li2017flame} for each selected timestamp. For each subject, we enforce consistent shape and static offset parameters across all timestamps and remove outliers based on tracking error statistics. This produces stable alignments between reconstructed geometry and parametric face models.

We preserve COLMAP geometric detail while aligning reconstructions into a shared canonical space. We construct canonical point clouds by transferring FLAME-tracking deformations to reconstructed points. Given tracked and canonical FLAME meshes, we compute per-vertex deformation vectors between corresponding surface points. Each COLMAP point receives the deformation of its nearest FLAME surface point, mapping reconstructed points into the FLAME canonical-pose, neutral-expression coordinate system at metric scale.

Using these canonicalized point clouds, we generate canonical maps that describe the canonical location of each pixel in the input images. These maps provide dense supervision for learning canonical correspondences, while the COLMAP depth maps provide supervision for detailed geometry reconstruction. Although depth supervision can be sparse in some regions, the strong geometric prior learned from DA3 enables reliable generalization to these areas.

\begin{figure*}[t]
\centering
\includegraphics[width=\textwidth]{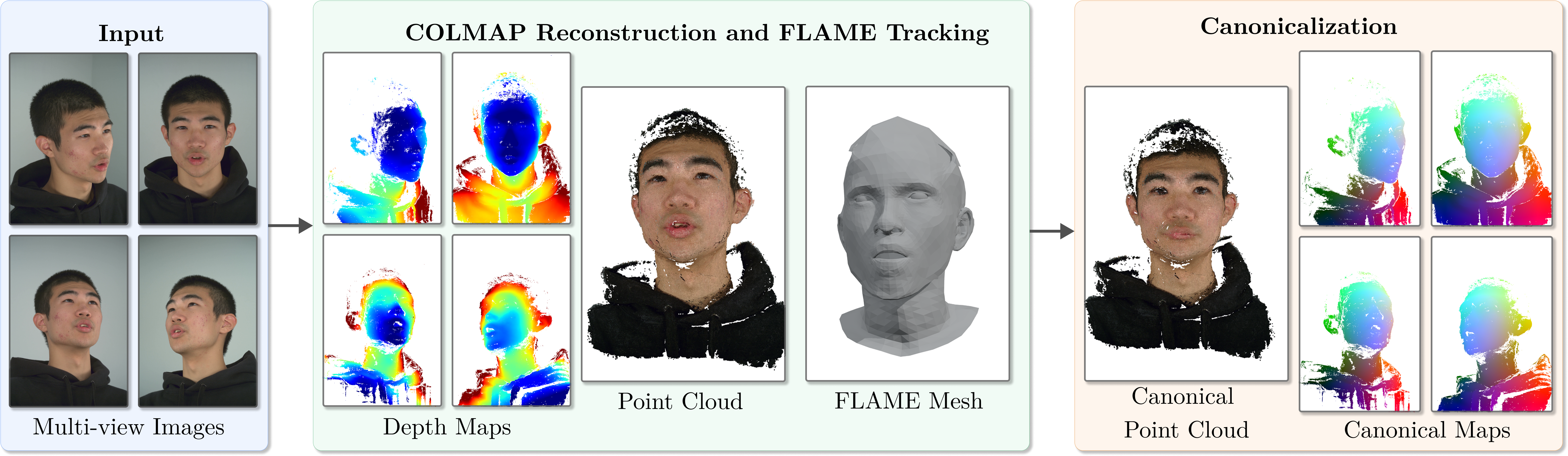}

\caption{
\textbf{Dataset creation.} We generate training supervision by combining multi-view reconstruction with parametric face tracking to produce depth maps and canonical facial maps. Although the parametric face model may not capture fine-scale geometric details, high-frequency information from COLMAP reconstruction is preserved in the canonical maps. This process provides geometrically consistent
supervision across viewpoints, expressions, and identities for training our model.
}
\label{fig:framework_dataset}

\end{figure*}

The resulting dataset provides multi-view RGB images, calibrated camera parameters, depth maps, and dense canonical maps across a wide range of identities, poses, and expressions. Because canonical maps are defined in a normalized FLAME coordinate system, correspondences across subjects and frames exhibit strong structural consistency, making them well-suited for supervised learning. Despite minor inaccuracies from parametric tracking, we observe that the network learns to compensate for small alignment errors during training. An overview of the dataset creation process is shown in \cref{fig:framework_dataset}.

\begin{figure*}[t]
\centering

\begin{tabular}{@{}ccccccccc@{}}

\makebox[0.102\textwidth][c]{\resizebox{!}{0.5em}{Pi3~\cite{wang2025pi}}} &
\makebox[0.102\textwidth][c]{\resizebox{!}{0.5em}{DA3~\cite{lin2025da3}}} &
\makebox[0.102\textwidth][c]{\resizebox{!}{0.5em}{VGGT~\cite{wang2025vggt}}} &
\makebox[0.102\textwidth][c]{\resizebox{!}{0.5em}{DAViD~\cite{saleh2025david}}} &
\makebox[0.102\textwidth][c]{\resizebox{!}{0.5em}{V-DPM~\cite{sucar2025vdpm}}} &
\makebox[0.102\textwidth][c]{\resizebox{!}{0.5em}{Sapiens~\cite{khirodkar2024sapiens}}} &
\makebox[0.102\textwidth][c]{\resizebox{!}{0.5em}{Ours}} &
\makebox[0.102\textwidth][c]{\resizebox{!}{0.5em}{RGB}} &
\makebox[0.102\textwidth][c]{\resizebox{!}{0.5em}{GT}} \\

\end{tabular}

\includegraphics[width=\textwidth]{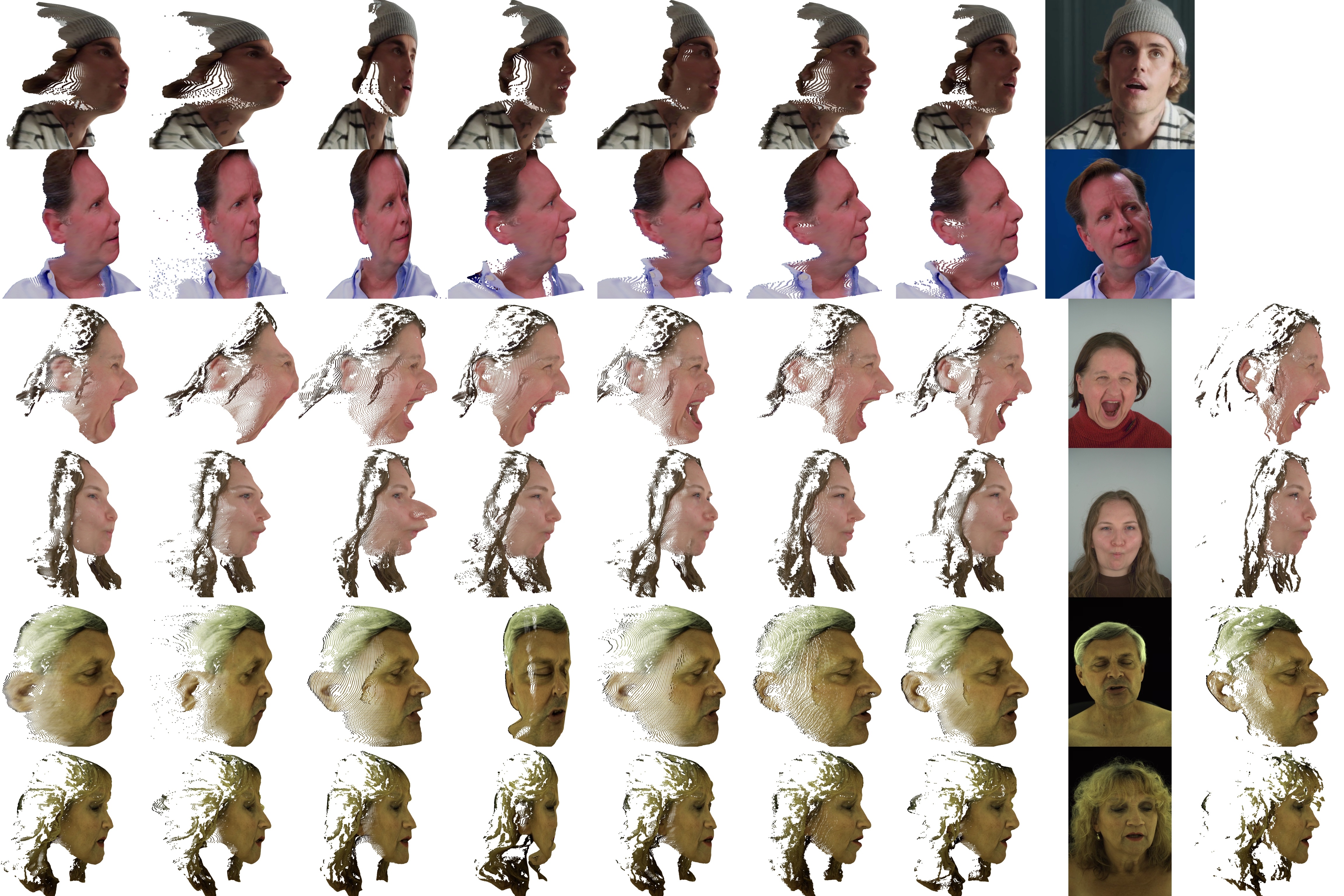}

\caption{
\textbf{4D reconstruction comparison on VFHQ, NeRSemble, and Ava-256.} COLMAP reconstructions are shown as pseudo ground truth for NeRSemble and Ava-256, while VFHQ does not provide ground-truth geometry. Our method produces more accurate and detailed reconstructions than recent approaches.
}
\label{fig:points_qualitative}

\end{figure*}

\section{Experiments}
\label{sec:experiments}

\subsection{Implementation Details}
\label{sec:experiments_implementation_details}

We train the model in two stages. First, our 1.2B-parameter model is pretrained on approximately 100k facial images from the DAViD dataset~\cite{saleh2025david} to learn geometric facial priors. Then, the main training stage is performed for 90 epochs using the AdamW optimizer~\cite{adamW2019} with a learning rate initialized at $2\times10^{-5}$ and decayed to $1\times10^{-8}$ using a cosine scheduler, together with gradient clipping of 1. Each epoch consists of 800 sampled batches containing up to 48 images, while each sequence contains between 2 and 16 images with gradient accumulation over three steps. Training is performed using bfloat16 precision with gradient checkpointing for efficiency. Following DA3~\cite{lin2025da3}, we train using multiple input resolutions with a base resolution of $504\times504$.

\subsection{Evaluation Protocol}
\label{sec:experiments_evaluation_protocol}

\begin{table}[t]
\caption{
\textbf{Monocular depth estimation on NeRSemble and Ava-256.}
Results are reported for both images and videos in metric scale as Image/Video. Best results are shown in bold. Values are $\times 10$.
}
\centering
\small
\setlength{\tabcolsep}{6pt}
\renewcommand{\arraystretch}{1.15}
\scalebox{1.02}{
\begin{tabular}{lcccc}
\toprule

&
\multicolumn{2}{c}{NeRSemble}
& \multicolumn{2}{c}{Ava-256} \\

\cmidrule(lr){2-3}
\cmidrule(lr){4-5}

\textbf{Setting}
& RMSE$\downarrow$ & AbsRel$\downarrow$
& RMSE$\downarrow$ & AbsRel$\downarrow$ \\

\midrule

Pi3~\cite{wang2025pi}
& 0.193/0.160 & 0.134/0.108
& 0.083/0.091 & 0.066/0.071 \\

DA3~\cite{lin2025da3}
& 0.162/0.127 & 0.100/0.085
& 0.148/0.116 & 0.129/0.100 \\

VGGT~\cite{wang2025vggt}
& 0.115/0.119 & 0.076/0.080
& 0.084/0.103 & 0.061/0.084 \\

Sapiens-1B~\cite{khirodkar2024sapiens}
& 0.112/0.112 & 0.065/0.065
& 0.079/0.079 & 0.059/0.059 \\

DAViD~\cite{saleh2025david}
& 0.110/0.110 & 0.061/0.061
& 0.182/0.182 & 0.160/0.160 \\

V-DPM~\cite{sucar2025vdpm}
& 0.102/0.104 & 0.061/0.062
& 0.090/0.097 & 0.070/0.075 \\

Sapiens-2B~\cite{khirodkar2024sapiens}
& 0.085/0.085 & 0.048/0.048
& 0.081/0.081 & 0.056/0.056 \\

\midrule

Ours
& \textbf{0.077/0.075} & \textbf{0.040/0.038}
& \textbf{0.067/0.065} & \textbf{0.048/0.048} \\

\bottomrule
\end{tabular}
}
\label{tab:combined_depth_results}
\end{table}

\begin{figure}[t]
\centering
\setlength{\tabcolsep}{1pt}

\begin{tabular}{cccccccc}

Base &
Support &
Single &
Multi &
Base &
Support &
Single &
Multi \\

\includegraphics[width=0.12\linewidth]{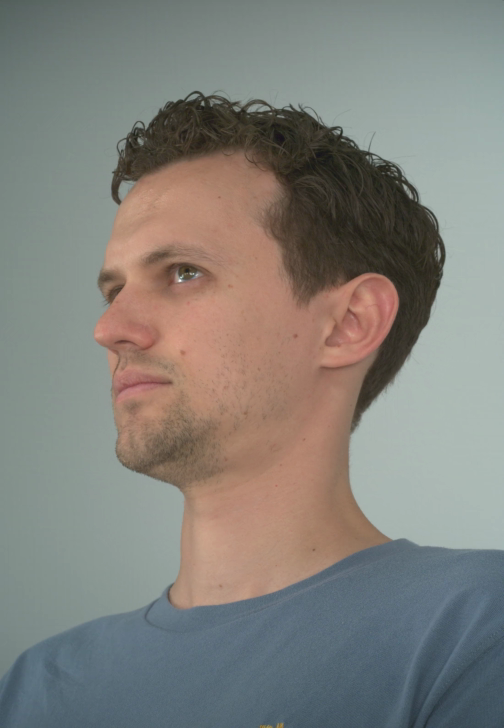} &
\includegraphics[width=0.12\linewidth]{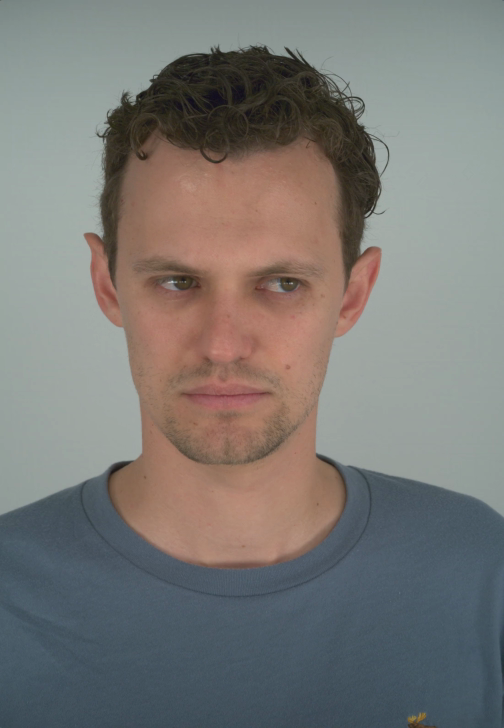} &
\includegraphics[width=0.12\linewidth]{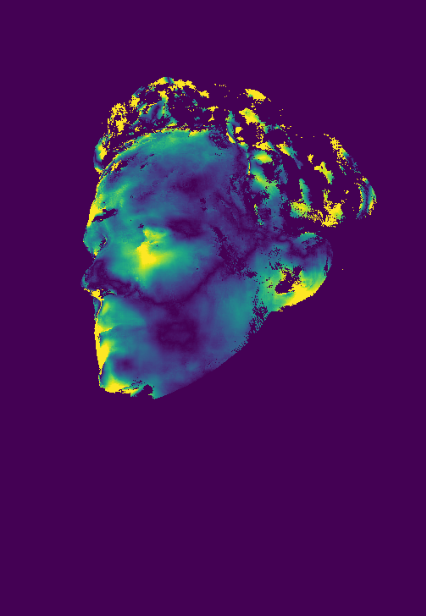} &
\includegraphics[width=0.12\linewidth]{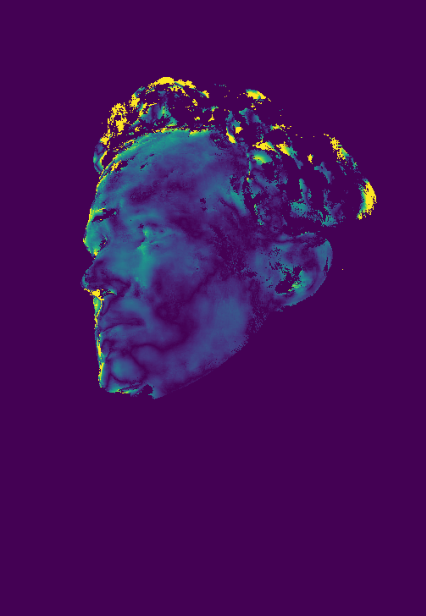} &

\includegraphics[width=0.12\linewidth]{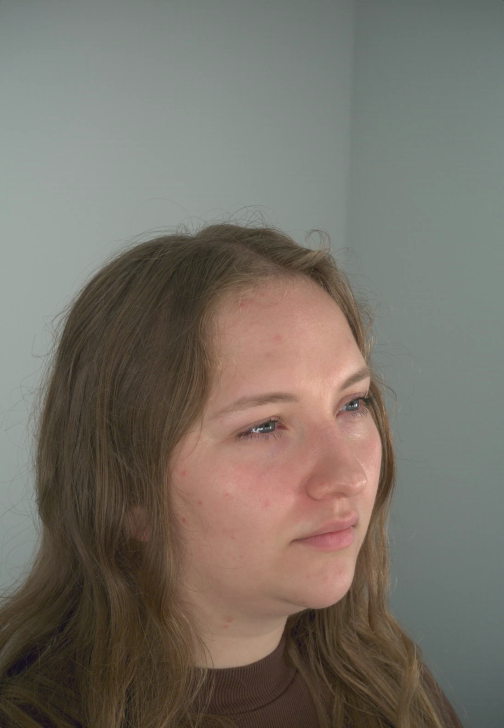} &
\includegraphics[width=0.12\linewidth]{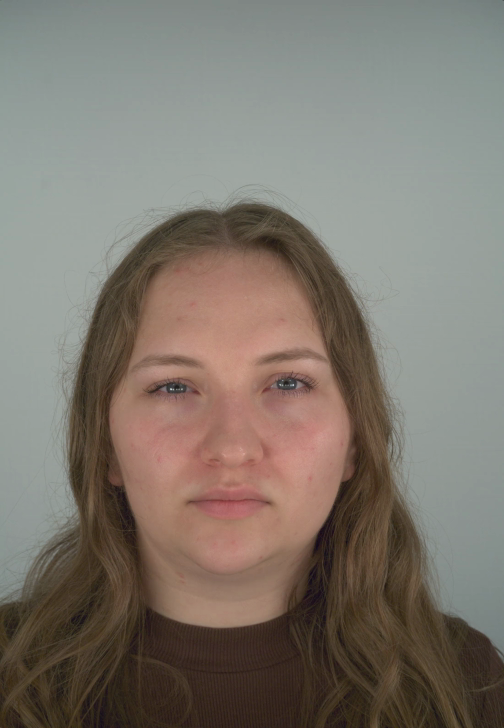} &
\includegraphics[width=0.12\linewidth]{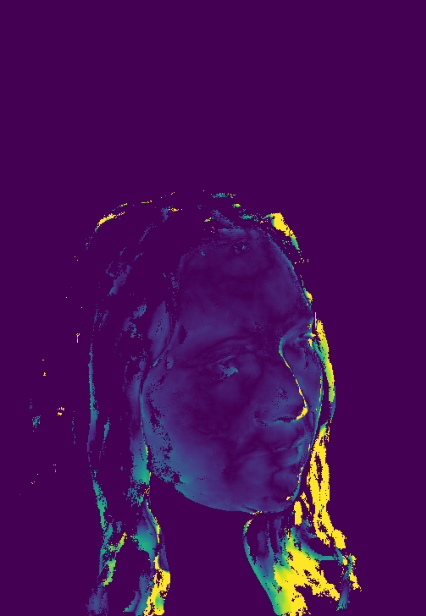} &
\includegraphics[width=0.12\linewidth]{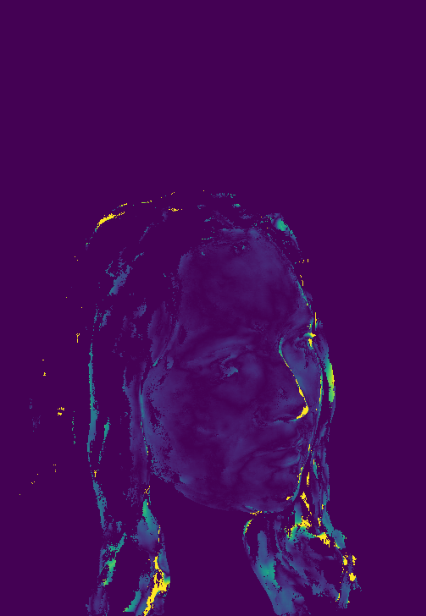}

\end{tabular}

\caption{
\textbf{Single-view vs multi-view depth prediction.} Darker colors indicate lower error. Multi-view input improves depth accuracy over monocular prediction.
}

\label{fig:single_multiview_error_maps}

\end{figure}

\textbf{Evaluation Datasets.}
We evaluate our method on various datasets to measure both reconstruction fidelity and generalization to in-the-wild data, including NeRSemble~\cite{kirschstein2023nersemble}, Ava-256~\cite{martinez2024codec}, VFHQ~\cite{xie2022vfhq}, and CelebV-HQ~\cite{zhu2022celebvhq}. From NeRSemble and Ava-256 we select five subjects each; NeRSemble subjects are not used during training and include diverse sequences, while Ava-256 samples are obtained by randomly selecting 40 images across all sequences. For VFHQ, we evaluate on all test videos, and for CelebV-HQ, we randomly select a subset of videos. For consistency across datasets, we use the frontal camera views.

\textbf{Evaluation Metrics.}
Depth accuracy is evaluated using the Root Mean Squared Error (RMSE) and Absolute Relative Error (AbsRel). Correspondence accuracy is measured using end-point error (EPE) in both 2D and 3D, where $P_i(t_j)$ and various temporal margins follow the definition in V-DPM~\cite{sucar2025vdpm}. We additionally report the percentage of correspondences within different pixel thresholds ($<n$px). Geometric consistency is evaluated using Forward-Backward Cycle Consistency Error (CCE) of correspondence predictions, while photometric alignment is assessed with Warping Photometric Error (WPE) after warping correspondences between frames. FLAME tracking accuracy is measured using the unidirectional L1 Chamfer distance (CD-L1) from ground-truth COLMAP points to the predicted mesh.

\begin{figure*}[t]
\centering

\begin{tabular}{@{}cccccc@{}}
\makebox[0.158\textwidth][c]{Base RGB} &
\makebox[0.158\textwidth][c]{Target RGB} &
\makebox[0.158\textwidth][c]{Tracks} &
\makebox[0.158\textwidth][c]{P3DMM~\cite{giebenhain2025pixel3dmm}} &
\makebox[0.158\textwidth][c]{V-DPM~\cite{sucar2025vdpm}} &
\makebox[0.158\textwidth][c]{Ours} \\
\end{tabular}

\includegraphics[width=\textwidth]{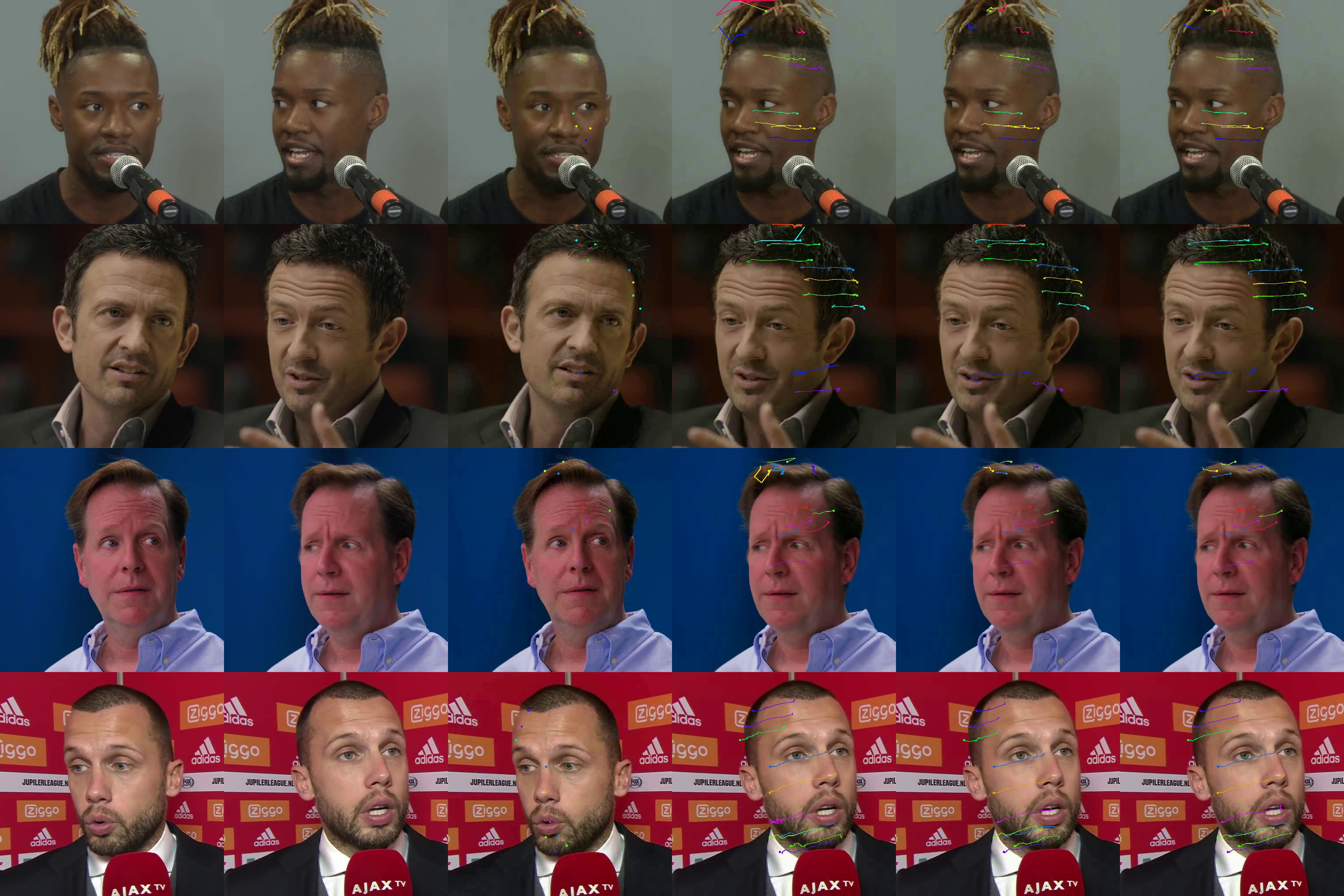}

\caption{
\textbf{2D tracking comparison on VFHQ.} Track points are defined in the base image and each method predicts trajectories to the target image that should end at the same facial locations. Our method produces more accurate and consistent correspondences than recent approaches.
}
\label{fig:2dtracking_qualitative}

\end{figure*}

\subsection{4D Reconstruction}
\label{sec:experiments_4d_reconstruction}

We evaluate our method on 4D facial reconstruction and compare against state-of-the-art approaches in qualitative and quantitative settings. For multi-image methods, we provide all input views and evaluate on the first 40 frames of each sequence, while monocular methods are applied independently to each frame.

\textbf{Qualitative Reconstruction.}
\cref{fig:points_qualitative} presents qualitative comparisons of reconstructed 3D facial geometry across methods. Our approach produces high-fidelity reconstructions with detailed facial geometry, whereas baseline methods often exhibit geometric artifacts or reduced detail.

\textbf{Depth Accuracy.}
Depth evaluation is reported in \cref{tab:combined_depth_results} for both image-based and video-based reconstruction settings. In the video-based setting, all frames of the input sequence are provided jointly as input. Our method achieves state-of-the-art performance across datasets despite using a smaller model than the largest Sapiens~\cite{khirodkar2024sapiens} variant and requiring significantly less training time.

\textbf{Effect of Multi-Image Prediction.}
To analyze the impact of multi-image inference on the NeRSemble dataset, \cref{fig:single_multiview_error_maps} compares depth error maps obtained using a single input image and using an additional supporting view. Incorporating supporting images clearly improves depth accuracy, particularly in side-view regions where monocular prediction is more ambiguous.

\begin{table}[t]
\caption{
\textbf{2D correspondence evaluation on NeRSemble.}
Dense correspondence accuracy is measured using EPE and percentage of predictions within pixel thresholds across temporal margins.
}
\centering
\setlength{\tabcolsep}{8pt}
\renewcommand{\arraystretch}{1.2}
\scalebox{0.68}{
\begin{tabular}{lcccccccc}
\toprule
& \multicolumn{4}{c}{\textbf{Margin = 2}} 
& \multicolumn{4}{c}{\textbf{Margin = 8}} \\
\cmidrule(lr){2-5} \cmidrule(lr){6-9}

Method 
& EPE(2D)$\downarrow$ & <3px$\uparrow$ & <5px$\uparrow$ & <10px$\uparrow$
& EPE(2D)$\downarrow$ & <3px$\uparrow$ & <5px$\uparrow$ & <10px$\uparrow$ \\

\midrule

P3DMM~\cite{giebenhain2025pixel3dmm} (w/o Hair)
& 3.089 & 0.629 & 0.873 & 0.986
& 3.971 & 0.508 & 0.768 & 0.955 \\

Ours (w/o Hair)
& \textbf{1.719} & \textbf{0.876} & \textbf{0.968} & \textbf{0.995}
& \textbf{2.314} & \textbf{0.769} & \textbf{0.926} & \textbf{0.990} \\

\midrule

P3DMM~\cite{giebenhain2025pixel3dmm} (w/ Hair)
& 5.550 & 0.565 & 0.797 & 0.927
& 6.597 & 0.446 & 0.686 & 0.883 \\

Ours (w/ Hair)
& \textbf{1.838} & \textbf{0.853} & \textbf{0.960} & \textbf{0.995}
& \textbf{2.470} & \textbf{0.741} & \textbf{0.913} & \textbf{0.987} \\

\bottomrule
\end{tabular}
}
\label{tab:2dcorrespondence_comparison}
\end{table}

\begin{table}[t]
\caption{
\textbf{Dense correspondence evaluation on VFHQ.}
Correspondence accuracy is evaluated using CCE and WPE at different temporal margins.
}
\centering
\setlength{\tabcolsep}{8pt}
\renewcommand{\arraystretch}{1.2}
\scalebox{0.74}{
\begin{tabular}{lcccccc}
\toprule

\multicolumn{7}{c}{\textbf{Margin = 5}} \\
\cmidrule(lr){1-7}

Method
& $CCE_{mean}\downarrow$
& $CCE_{median}\downarrow$
& $CCE_{<2px}\uparrow$
& $WPE_{L1}\downarrow$
& $WPE_{Grad}\downarrow$
& $WPE_{SSIM}\uparrow$ \\

\midrule

P3DMM~\cite{giebenhain2025pixel3dmm}
& 2.472 & 1.146 & 0.702 & 0.032 & 0.016 & 0.798 \\

V-DPM~\cite{sucar2025vdpm}
& 0.797 & 0.610 & 0.885 & 0.026 & 0.015 & 0.855 \\

Ours
& \textbf{0.398} & \textbf{0.007} & \textbf{0.961}
& \textbf{0.023} & \textbf{0.014} & \textbf{0.881} \\

\midrule

\multicolumn{7}{c}{\textbf{Margin = 20}} \\
\cmidrule(lr){1-7}

Method
& $CCE_{mean}\downarrow$
& $CCE_{median}\downarrow$
& $CCE_{<2px}\uparrow$
& $WPE_{L1}\downarrow$
& $WPE_{Grad}\downarrow$
& $WPE_{SSIM}\uparrow$ \\

\midrule

P3DMM~\cite{giebenhain2025pixel3dmm}
& 2.797 & 1.146 & 0.673 & 0.043 & 0.017 & 0.740 \\

V-DPM~\cite{sucar2025vdpm}
& 1.348 & 1.054 & 0.756 & 0.040 & 0.017 & 0.769 \\

Ours
& \textbf{0.774} & \textbf{0.069} & \textbf{0.909}
& \textbf{0.034} & \textbf{0.016} & \textbf{0.810} \\

\bottomrule
\end{tabular}
}
\label{tab:vfhq_quantitative_comparison}
\end{table}

\begin{table}[t]
\caption{
\textbf{3D correspondence evaluation on NeRSemble.}
3D correspondence accuracy is measured using EPE across temporal margins for reconstruction and tracking.
}
\centering
\setlength{\tabcolsep}{7pt}
\renewcommand{\arraystretch}{1.2}
\scalebox{0.69}{
\begin{tabular}{lccccccccc}
\toprule

& \multicolumn{4}{c}{\textbf{Margin = 2}}
& \multicolumn{4}{c}{\textbf{Margin = 8}}
& \textbf{Margin = 1--10} \\

\cmidrule(lr){2-5}
\cmidrule(lr){6-9}
\cmidrule(lr){10-10}

Method
& $P_0(t_0)$$\downarrow$
& $P_0(t_1)$$\downarrow$
& $P_1(t_0)$$\downarrow$
& $P_1(t_1)$$\downarrow$
& $P_0(t_0)$$\downarrow$
& $P_0(t_1)$$\downarrow$
& $P_1(t_0)$$\downarrow$
& $P_1(t_1)$$\downarrow$
& EPE$\downarrow$ \\

\midrule

V-DPM~\cite{sucar2025vdpm}
& 0.014 & 0.015 & 0.014 & 0.014
& 0.014 & 0.016 & 0.014 & 0.015
& 0.015 \\

Ours
& \textbf{0.004} & \textbf{0.004} & \textbf{0.004} & \textbf{0.004}
& \textbf{0.004} & \textbf{0.005} & \textbf{0.005} & \textbf{0.004}
& \textbf{0.005} \\

\bottomrule
\end{tabular}
}
\label{tab:3dcorrespondence_combined}
\end{table}

\subsection{2D Tracking}
\label{sec:experiments_2d_tracking}

We evaluate dense 2D tracking accuracy on both controlled and in-the-wild datasets and compare against state-of-the-art facial correspondence methods.

\textbf{Qualitative Tracking.}
\cref{fig:2dtracking_qualitative} shows qualitative comparisons of 2D correspondences on the VFHQ~\cite{xie2022vfhq} dataset. We visualize tracks between a reference image and a target frame for different methods. Our approach produces more consistent and accurate tracks, particularly in regions with large motion and fine structures. P3DMM~\cite{giebenhain2025pixel3dmm} fails to track points reliably in hair regions, while V-DPM~\cite{sucar2025vdpm} frequently predicts inaccurate locations under large motions. In contrast, our method maintains stable correspondences across challenging frames.

\textbf{Quantitative Correspondence Accuracy.}
We report quantitative tracking accuracy on the NeRSemble~\cite{kirschstein2023nersemble} dataset using ground-truth correspondences in \cref{tab:2dcorrespondence_comparison}. Our method consistently outperforms P3DMM across different temporal intervals and for both face-only and full head regions including hair.

\textbf{Cycle and Photometric Consistency.}
Additional quantitative results on the VFHQ dataset are reported in \cref{tab:vfhq_quantitative_comparison}, where we compare against P3DMM and V-DPM using CCE and WPE. Our method achieves the best performance.

\begin{figure*}[t]
\centering

\begin{tabular}{@{}ccc@{}}
\makebox[0.32\textwidth][c]{V-DPM~\cite{sucar2025vdpm}} &
\makebox[0.32\textwidth][c]{Ours} &
\makebox[0.32\textwidth][c]{RGB} \\
\end{tabular}

\includegraphics[width=\textwidth]{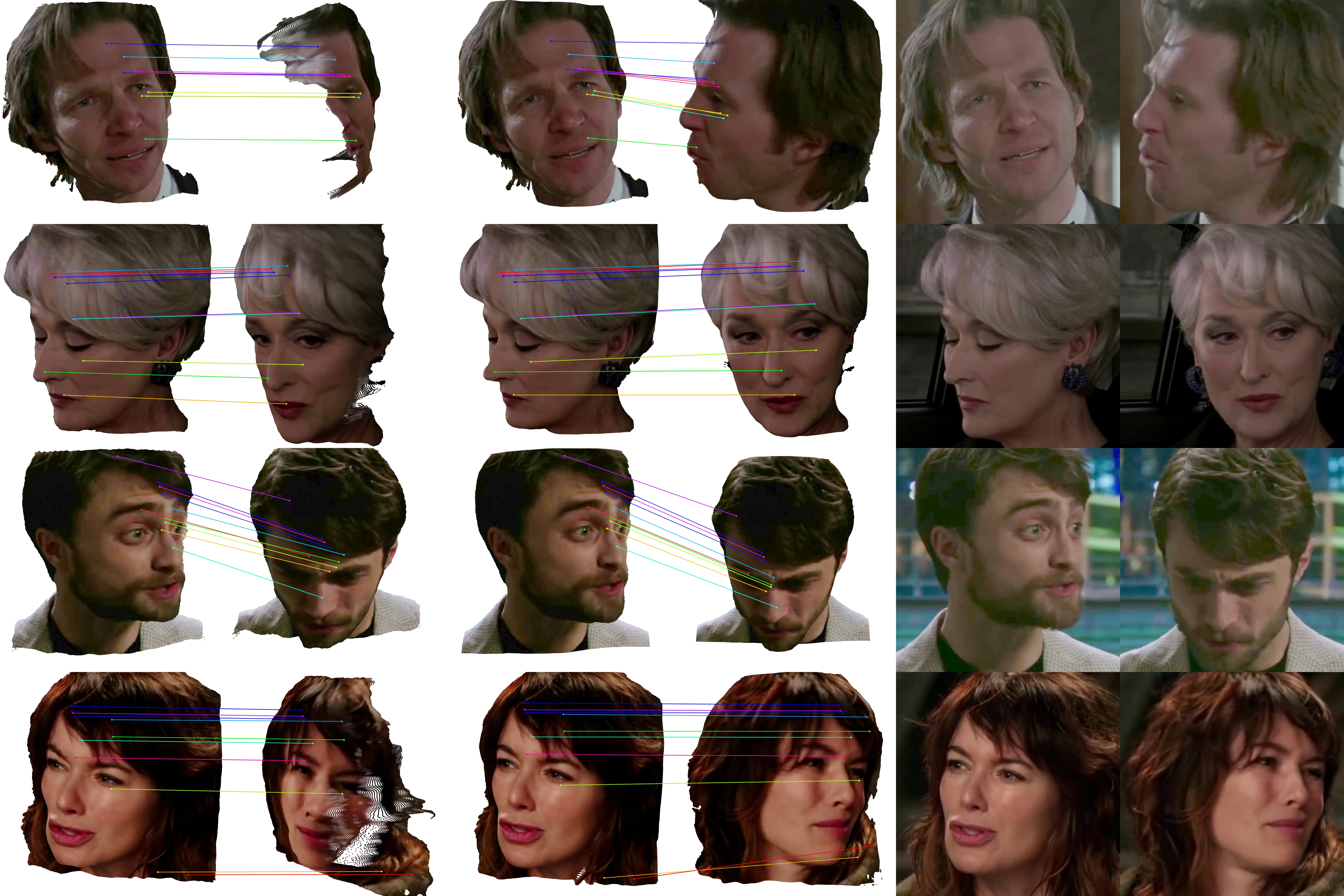}

\caption{
\textbf{4D reconstruction and tracking comparison on CelebV-HQ.} Both input images are reconstructed and correspondences are visualized with track lines between frames. Camera locations are fixed in the first image across methods for fair comparison. 
}
\label{fig:tracking_qualitative_celebvhq}

\end{figure*}

\subsection{Point Tracking}
\label{sec:experiments_point_tracking}

We evaluate dense 3D point tracking accuracy and efficiency of our method.

\textbf{Qualitative Tracking.}
\cref{fig:tracking_qualitative_celebvhq} presents qualitative comparisons with V-DPM on the CelebV-HQ~\cite{zhu2022celebvhq} dataset. Our method produces more accurate 4D reconstructions and more stable point trajectories over time, while V-DPM exhibits larger geometric inconsistencies and tracking errors. The results demonstrate that our canonical-space formulation leads to more reliable correspondences. 

\textbf{Quantitative Tracking Accuracy.}
\cref{tab:3dcorrespondence_combined} reports tracking accuracy across temporal margins. Our method achieves lower errors than V-DPM for both short- and long-range correspondences.

\textbf{Efficiency Analysis.}
We compare inference efficiency in \cref{tab:runtime_memory}, reporting runtime and GPU memory usage measured on 40 images and the maximum number of images that fit on a single GPU (all experiments at $518\times518$). Our method achieves favorable efficiency while supporting larger batch sizes.

\textbf{Temporal Correspondence Accuracy.}
\cref{fig:comparison_correspondence_temporal} shows correspondence error maps across temporal intervals. Our method produces consistently lower errors and cleaner correspondence structures compared to competing approaches, further demonstrating the accuracy of our tracking formulation.

\begin{table}[t]
\centering

\begin{minipage}{0.57\linewidth}
\centering
\captionof{table}{
\textbf{Runtime and memory efficiency comparison.}
Inference runtime and GPU memory usage measured on a single GPU (80GB).
}
\label{tab:runtime_memory}

\small
\setlength{\tabcolsep}{6pt}
\renewcommand{\arraystretch}{1.15}
\scalebox{0.57}{
\begin{tabular}{lccc}
\toprule

\textbf{Method} 
& \textbf{Runtime (s)$\downarrow$} 
& \textbf{Peak Memory (GB)$\downarrow$}
& \textbf{Max Images/GPU$\uparrow$} \\

\midrule

V-DPM~\cite{sucar2025vdpm}
& 160 
& 40
& 74 \\

Ours
& \textbf{5} 
& \textbf{19}
& \textbf{470} \\

\bottomrule
\end{tabular}
}
\end{minipage}
\hfill
\begin{minipage}{0.39\linewidth}
\centering

\captionof{table}{\textbf{FLAME tracking accuracy on NeRSemble.} Values are $\times 100$.}
\label{tab:flame_tracking_metric}

\small
\setlength{\tabcolsep}{6pt}
\renewcommand{\arraystretch}{1.15}
\scalebox{0.65}{
\begin{tabular}{lc}
\toprule
\textbf{Method} & \textbf{CD-L1$\downarrow$} \\
\midrule
P3DMM~\cite{giebenhain2025pixel3dmm} & 0.238 \\
Ours & \textbf{0.195} \\
\bottomrule
\end{tabular}
}
\end{minipage}

\end{table}

\begin{figure}[t]
\centering
\setlength{\tabcolsep}{1pt}

\textbf{(a)} Correspondence Error Comparison

\begin{tabular}{cccccccc}

\makebox[0.12\textwidth][c]{Base} &
\makebox[0.12\textwidth][c]{Target} &
\makebox[0.12\textwidth][c]{P3DMM} &
\makebox[0.12\textwidth][c]{Ours} &

\makebox[0.12\textwidth][c]{Base} &
\makebox[0.12\textwidth][c]{Target} &
\makebox[0.12\textwidth][c]{P3DMM} &
\makebox[0.12\textwidth][c]{Ours} \\

\includegraphics[width=0.112\textwidth]{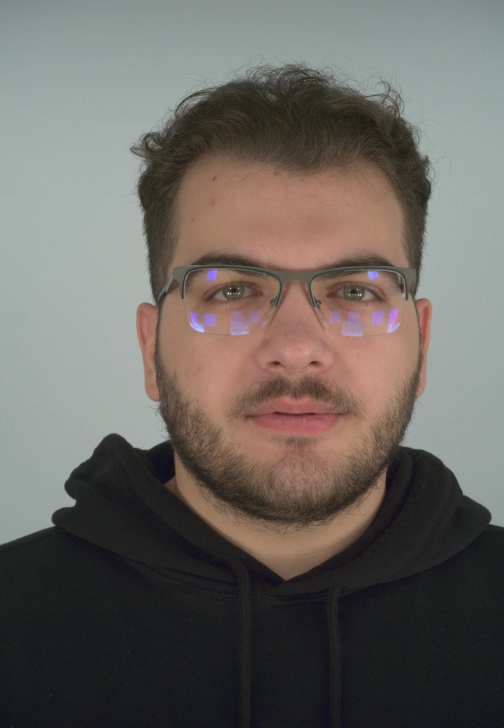} &
\includegraphics[width=0.112\textwidth]{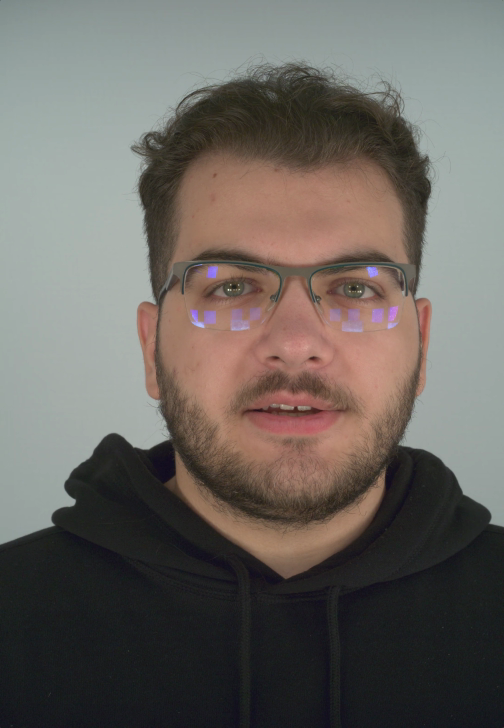} &
\includegraphics[width=0.112\textwidth]{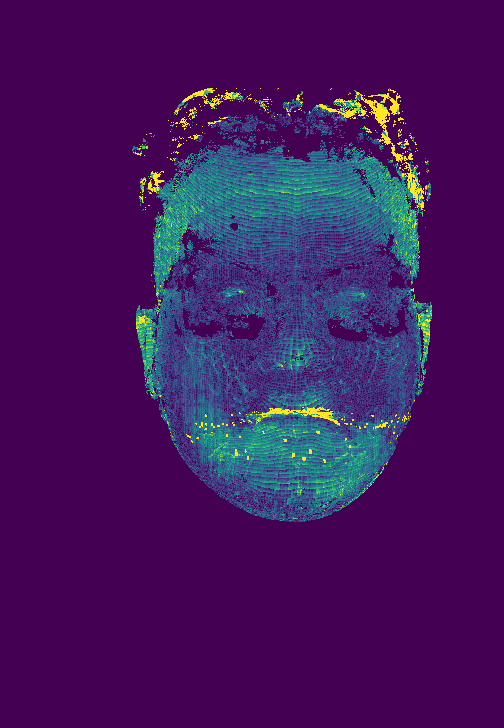} &
\includegraphics[width=0.112\textwidth]{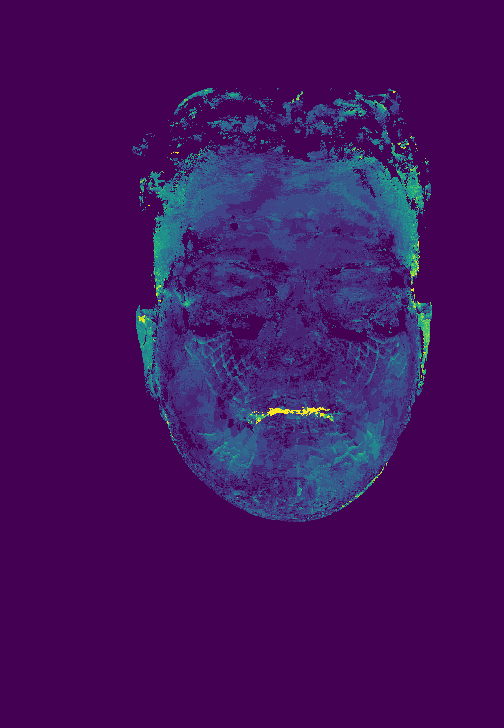} &

\includegraphics[width=0.112\textwidth]{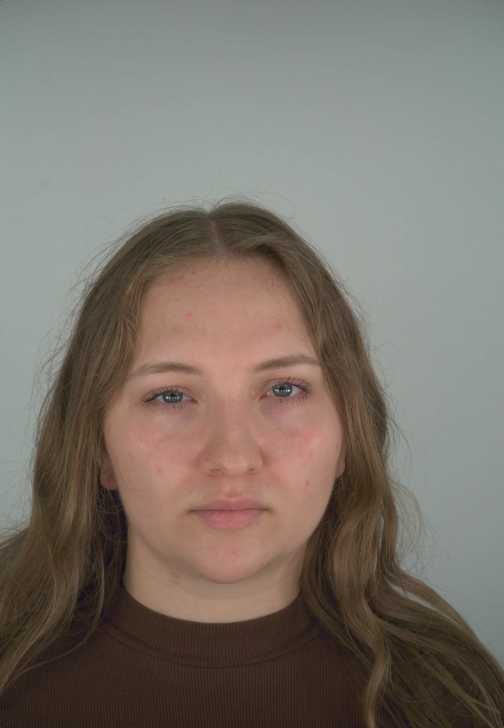} &
\includegraphics[width=0.112\textwidth]{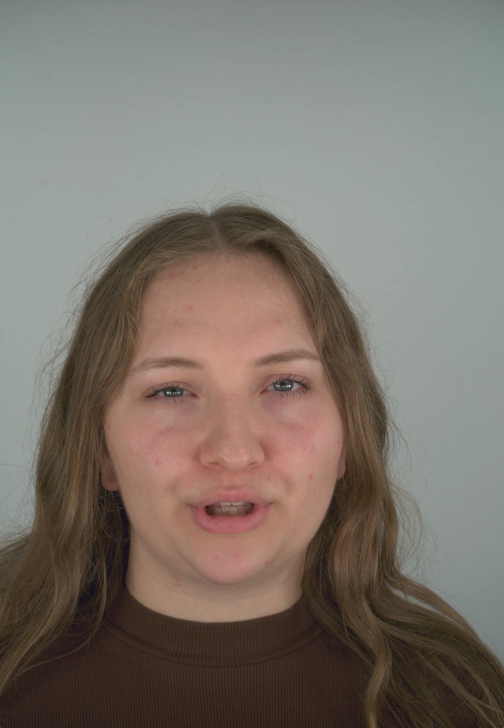} &
\includegraphics[width=0.112\textwidth]{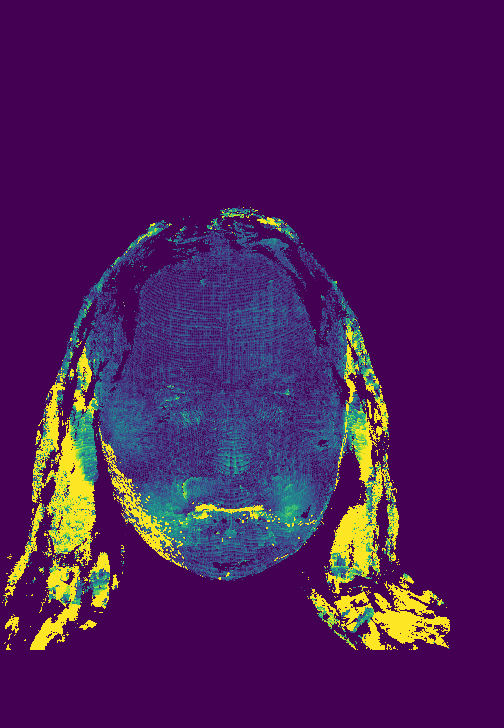} &
\includegraphics[width=0.112\textwidth]{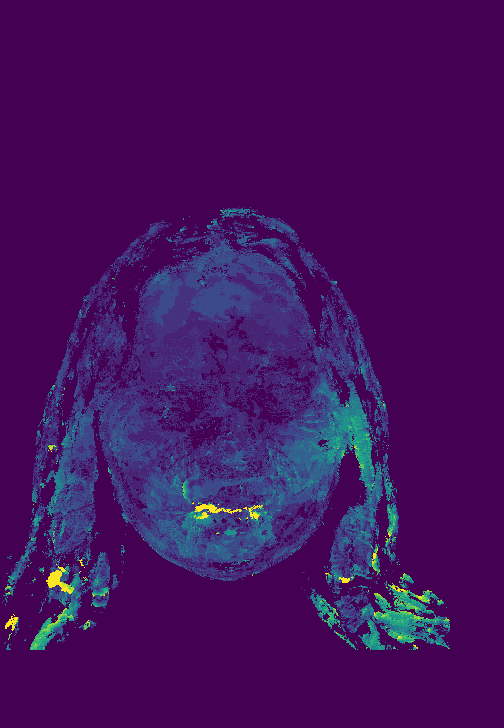} \\

\end{tabular}

\textbf{(b)} Temporal Prediction Errors

\begin{tabular}{cc|cccc|cccc}

\multicolumn{2}{c}{} &
\multicolumn{4}{c}{V-DPM} &
\multicolumn{4}{c}{Ours} \\

Base &
Target &
$P_0(t_0)$ & $P_0(t_1)$ & $P_1(t_0)$ & $P_1(t_1)$ &
$P_0(t_0)$ & $P_0(t_1)$ & $P_1(t_0)$ & $P_1(t_1)$ \\

\includegraphics[width=0.094\textwidth]{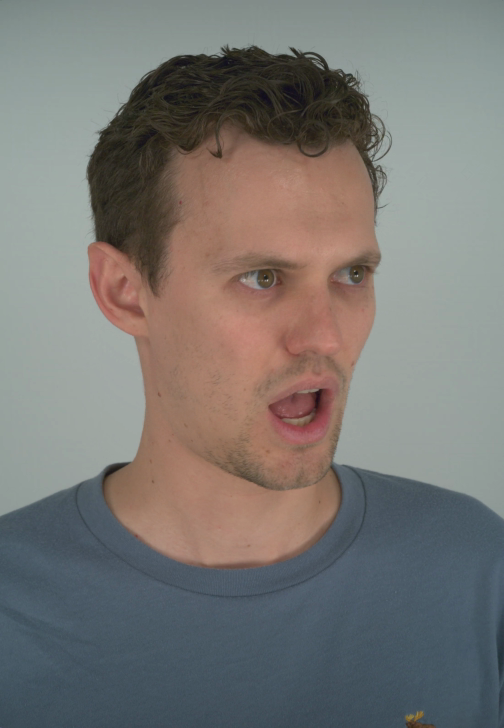} &
\includegraphics[width=0.094\textwidth]{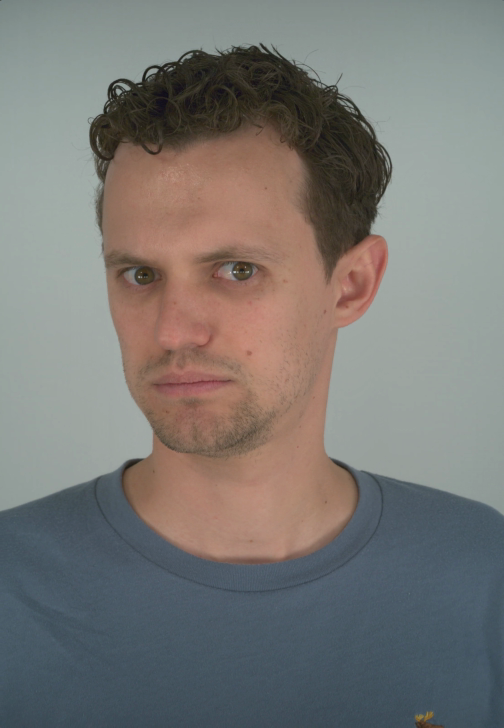} &

\includegraphics[width=0.094\textwidth]{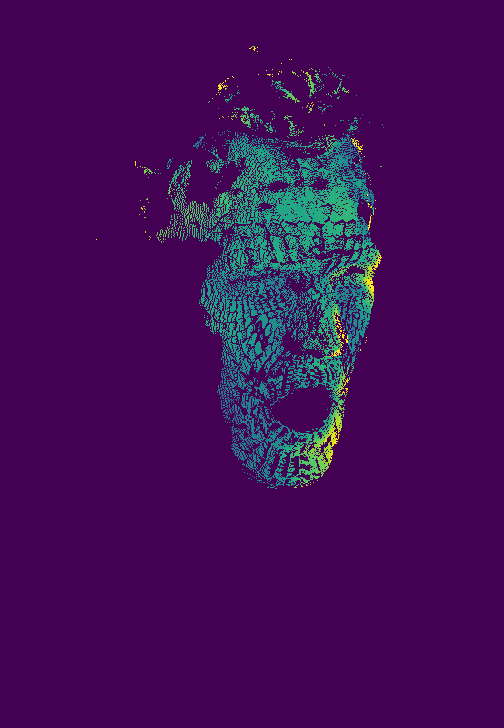} &
\includegraphics[width=0.094\textwidth]{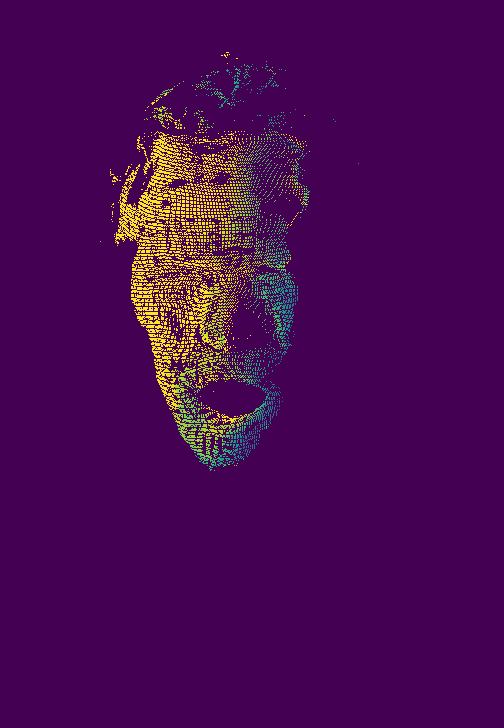} &
\includegraphics[width=0.094\textwidth]{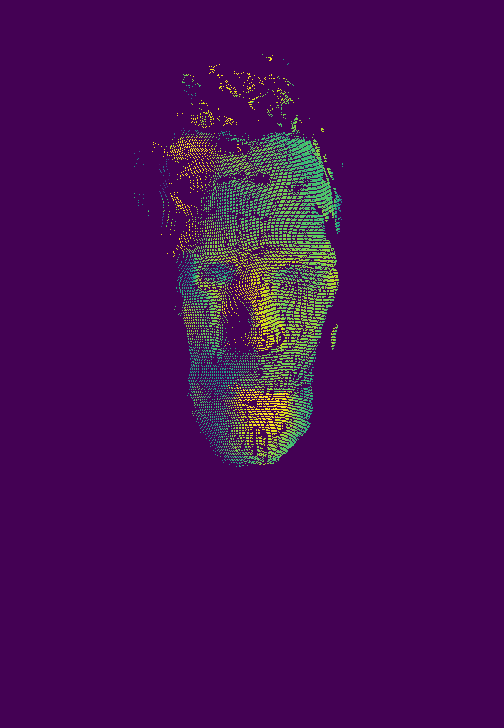} &
\includegraphics[width=0.094\textwidth]{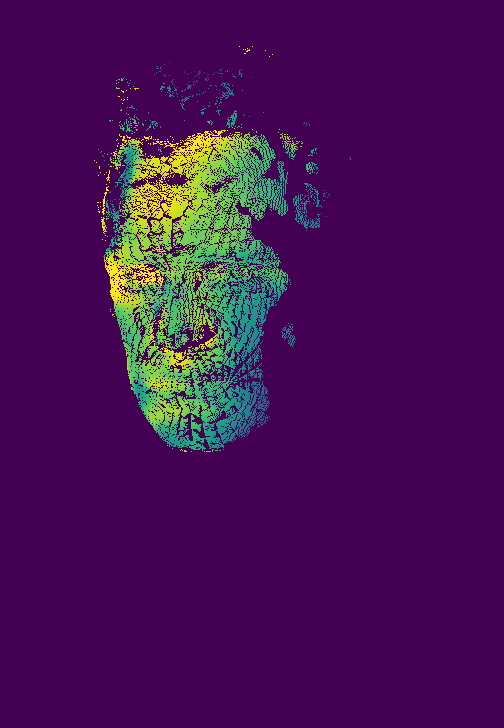} &

\includegraphics[width=0.094\textwidth]{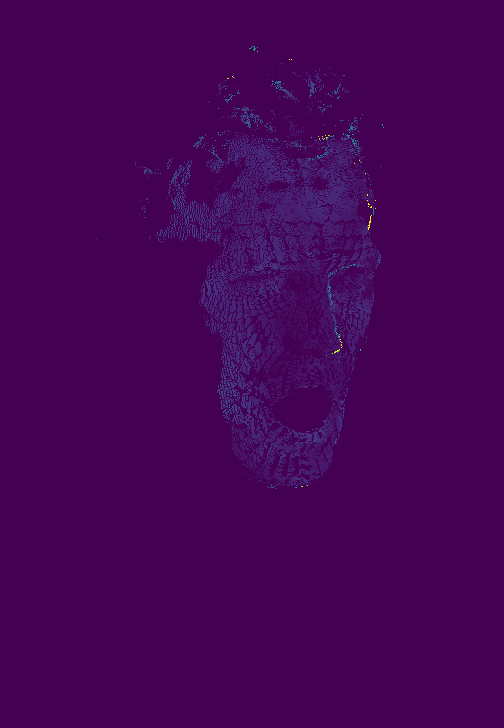} &
\includegraphics[width=0.094\textwidth]{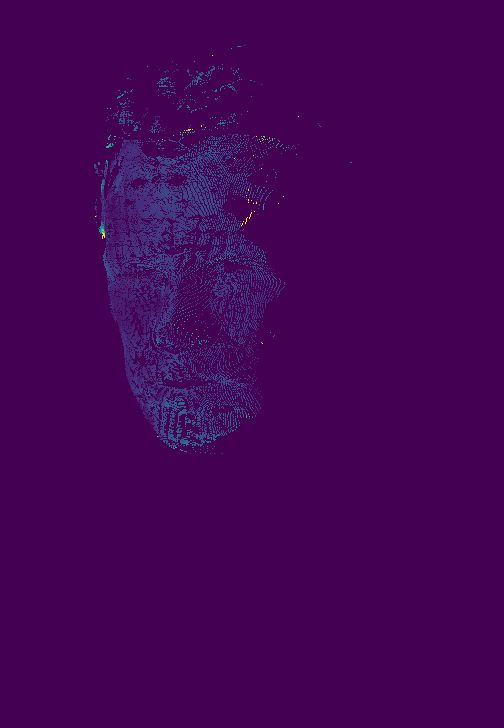} &
\includegraphics[width=0.094\textwidth]{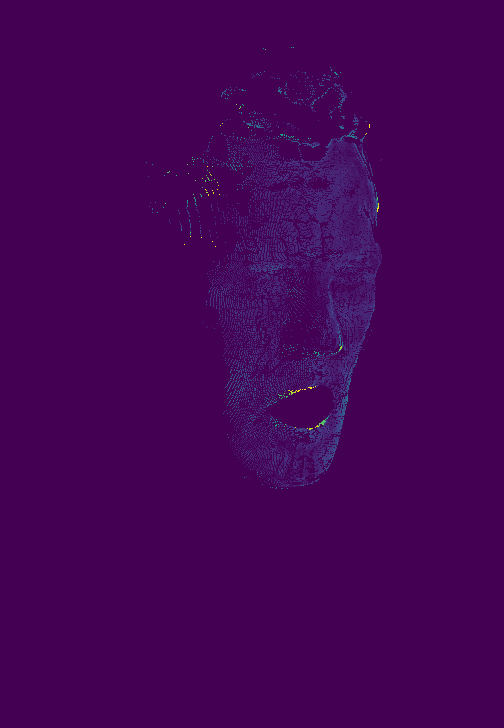} &
\includegraphics[width=0.094\textwidth]{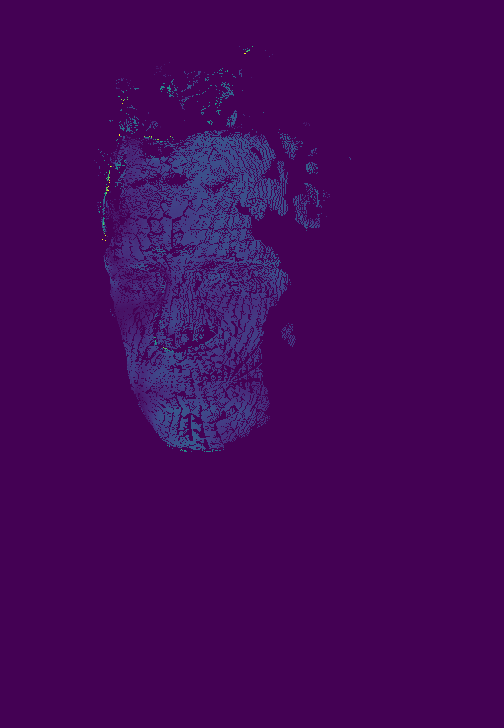} \\

\end{tabular}

\caption{
\textbf{Correspondence and temporal prediction errors on NeRSemble.}
(a) Pixel-wise correspondence errors for P3DMM~\cite{giebenhain2025pixel3dmm} and our method. 
(b) Temporal prediction errors compared to V-DPM~\cite{sucar2025vdpm}.
}

\label{fig:comparison_correspondence_temporal}
\end{figure}

\subsection{FLAME Tracking}
\label{sec:experiments_flame_tracking}

We evaluate monocular FLAME tracking accuracy when constrained by predictions from P3DMM and our method. As shown in \cref{tab:flame_tracking_metric}, our predictions lead to more accurate face tracking than constraints derived from P3DMM.

\subsection{Ablation Studies}
\label{sec:experiments_ablation_studies}

We conduct ablation studies on the NeRSemble dataset to analyze the effects of our training strategy and canonical correspondence formulation, as shown in \cref{tab:ablation}. Starting from DA3 as the baseline, we observe limited performance, particularly for multi-view depth estimation. Pretraining on DAViD improves monocular depth accuracy thanks to stronger facial priors, but does not generalize well to multi-view depth prediction.

We further evaluate two simplified training variants. Monocular Training, where the camera viewpoint is fixed within a forward pass while the timestamp varies, yields strong correspondence and camera estimation accuracy but leads to weaker multi-view depth predictions. In contrast, Static Training, where the timestamp is fixed while the camera viewpoint varies, improves multi-view depth estimation but degrades correspondence and camera estimation due to the lack of temporal variation.

We also evaluate an alternative correspondence formulation (Ours (Motion Pred)) that predicts motion from the current timestamp to the canonical space. This formulation performs significantly worse in correspondence prediction, suggesting that directly learning motion mappings is less stable.

Our final formulation samples both cameras and timestamps during training and predicts canonical maps instead of motion fields. As shown in \cref{tab:ablation}, this design achieves consistently strong performance across all metrics, producing accurate monocular and multi-view depth predictions while maintaining reliable correspondence estimation.

\begin{table}[t]
\caption{
\textbf{Ablation study.} Component analysis using depth (AbsRel), correspondence (EPE(2D)), and camera (Camera Rot ($^\circ$)) metrics. Best results in \textbf{bold}, second-best \underline{underlined}. AbsRel values are $\times 10$. The full design achieves consistently strong performance across all metrics, producing accurate monocular and multi-view depth predictions while maintaining reliable correspondence estimation.
}
\centering
\small
\setlength{\tabcolsep}{7pt}
\renewcommand{\arraystretch}{1.15}
\scalebox{0.88}{
\begin{tabular}{lccccc}
\toprule

\textbf{Method}
& AbsRel (Monocular)$\downarrow$
& EPE$\downarrow$
& AbsRel (16 views)$\downarrow$ 
& Camera Rot $\downarrow$\\

\midrule

DA3 (Baseline)
& 0.085
& -
& 0.076
& - \\

DAViD Pretrained
& 0.061
& -
& \underline{0.033}
& - \\

Monocular Training
& 0.054
& \textbf{3.031}
& 0.064
& \textbf{0.038} \\

Static Training
& \textbf{0.050}
& 3.864
& \textbf{0.015}
& 2.102 \\

Ours (Motion Pred)
& -
& 6.210
& -
& - \\

Ours 
& \underline{0.053}
& \underline{3.271}
& \textbf{0.015}
& \underline{0.054} \\

\bottomrule
\end{tabular}
}
\label{tab:ablation}
\end{table}

\section{Limitations and Future Work}
\label{sec:discussion}

Despite achieving strong performance on 4D facial reconstruction and tracking, our method has several limitations. The model is specialized for faces and relies on learned facial priors, which limits generalization to non-face scenes, and canonicalization of nearby objects such as microphones, hands, or accessories is often unreliable. Reconstruction quality can also degrade under strong occlusions, extreme viewpoints, or limited facial visibility where geometric cues are insufficient. Extending canonical map prediction beyond faces and improving robustness in challenging real-world scenarios are promising directions for future work. Another promising direction is integrating canonical prediction with generative or neural rendering models to enable controllable facial animation and avatar creation from monocular video.

\section{Conclusion}
\label{sec:conclusion}

We present Face Anything, a unified method for high-fidelity 4D facial reconstruction and dense tracking from image sequences. Our method jointly estimates depth and canonical facial coordinates, enabling temporally consistent reconstruction and reliable correspondences within a single feed-forward model. The proposed canonical representation simplifies correspondence learning and allows efficient multi-frame reconstruction and tracking without explicit motion modeling. To support supervised learning of canonical correspondences, we construct a dataset based on NeRSemble with multi-view geometry and canonical alignment. Extensive experiments demonstrate state-of-the-art performance across face depth estimation, 4D reconstruction, and dense tracking benchmarks while improving efficiency over prior methods. These results highlight canonical map prediction as an effective representation for dynamic facial understanding and a promising direction for future spatiotemporal reconstruction methods. 

\clearpage
\textit{Acknowledgments} This work was supported by the ERC Consolidator Grant Gen3D (101171131) of Matthias Nie{\ss}ner. We thank Angela Dai for the video voice-over.

\bibliographystyle{splncs04}
\bibliography{main}

\clearpage
\appendix
\section*{Appendix}
\addcontentsline{toc}{section}{Appendix}
\input{supplementary}

\end{document}

%% file: supplementary.tex
\section{Supplementary Video}
\label{sec:supplementary_video}

We highly recommend watching our supplementary video, which presents additional qualitative results of our method. The video demonstrates \textbf{4D facial reconstruction with point tracking}, illustrating the temporal consistency of the reconstructed geometry across frames.

We further provide orbiting camera visualizations of the reconstructed sequences to better showcase the recovered dynamic facial geometry from novel viewpoints.

Additionally, the video includes comparisons of 2D correspondence tracking between our method and baseline approaches. These visualizations highlight the accuracy and temporal stability of the correspondences estimated by our approach.

\section{Additional Implementation Details}
\label{sec:additional_implementation_details}

\subsection{Training Details}

During training, we apply a set of photometric augmentations to improve robustness to appearance variations. Specifically, we employ color jittering with brightness, contrast, and saturation factors of $0.5$, and hue variation of $0.1$. The color jitter augmentation is applied with probability $0.9$. In addition, grayscale augmentation is enabled during training.

To maintain photometric consistency across correlated inputs, we apply \textit{co-jittering}, where identical color perturbations are applied jointly across frames. This helps preserve relative color relationships between views while still providing appearance variation.

Training images are sampled from a predefined set of resolutions to improve robustness to scale changes. The set of resolutions used during training is:

\begin{itemize}
\item $(504, 504)$
\item $(378, 504)$
\item $(336, 504)$
\item $(280, 504)$
\item $(504, 336)$
\item $(504, 756)$
\item $(504, 672)$
\end{itemize}

For numerical stability, we normalize the ground-truth 3D world coordinates such that the mean $\ell_2$ norm of valid ground-truth points equals $1$. This normalization ensures a consistent geometric scale across different subjects and sequences.

Canonical coordinate maps are defined in the FLAME~\cite{li2017flame} coordinate system, where the origin is located at the center of the face. Using this canonical representation allows consistent alignment across subjects and facilitates stable learning of facial geometry.

\subsection{Dataset Details}

\textbf{NeRSemble.}
For evaluation on the NeRSemble~\cite{kirschstein2023nersemble} dataset, we use the following five test subjects:

\begin{itemize}
\item 043
\item 128
\item 236
\item 306
\item 474
\end{itemize}

To evaluate different aspects of our method, we select different sequences for different tasks in order to increase evaluation diversity.

For \textbf{depth and reconstruction evaluation}, we use the following sequences:

\begin{itemize}
\item 043\_SEN-01-cramp\_small\_danger
\item 128\_EMO-2-surprise+fear
\item 236\_EMO-1-shout+laugh
\item 306\_EXP-4-lips
\item 474\_EXP-3-cheeks+nose
\end{itemize}

For \textbf{tracking evaluation}, we use the following sequences:

\begin{itemize}
\item 043\_SEN-02-same\_phrase\_thirty\_times
\item 128\_EXP-9-jaw-2
\item 236\_EMO-4-disgust+happy
\item 306\_EXP-1-head
\item 474\_SEN-06-problems\_wise\_chief
\end{itemize}

Using different sequences across tasks ensures that the evaluation covers a wide range of expressions, motions, and speaking patterns.

\textbf{Ava-256.}
For evaluation on the Ava-256~\cite{martinez2024codec} dataset, we use the following subjects:

\begin{itemize}
\item 20210810--1306--FXN596
\item 20210817--0900--NRE683
\item 20210818--1332--CDR970
\item 20210819--0903--DOT682
\item 20210827--0906--KDA058
\end{itemize}

For all reported metrics, we restrict the evaluation to the facial region. Specifically, we use Facer~\cite{zheng2022farl} masks that include both the facial area and hair to define the evaluation region.

\textbf{FLAME prior.}
Our canonical supervision is not purely derived from FLAME~\cite{li2017flame}. Instead, FLAME is used only for coarse alignment across subjects, while high-frequency geometric details are obtained from COLMAP~\cite{schoenberger2016sfm} reconstructions. As shown in \cref{fig:flame_limitation}, this supervision preserves structures that are not represented by the FLAME topology, including eyeglasses, long hair, accessories, and mouth interiors. Consequently, the model learns high-fidelity visible-surface geometry that goes beyond the FLAME prior.

\begin{figure}[t]
\centering

\hspace{-10mm} Input Image \hspace{20mm} Point Maps \hspace{26mm}  Canonical Maps

\includegraphics[width=0.98\textwidth]{experiments/additionals/flame_limitation_addressing/final.jpg}\\[-1mm]

\caption{
\textbf{Canonicalization visualization.} Example input frames with the corresponding point maps and canonical maps used for supervision. The canonical maps retain consistent facial coordinates while preserving visible non-FLAME structures, including hair, accessories, and mouth interiors.
}
\label{fig:flame_limitation}

\end{figure}

\textbf{COLMAP tuning.}
We tune the COLMAP~\cite{schoenberger2016sfm} reconstruction parameters on a 20-subject validation split to improve coverage of challenging facial regions. Increasing \path{PatchMatchStereo.num_samples} from $15$ to $50$ and \path{PatchMatchStereo.window_radius} from $5$ to $9$ improves facial coverage from $77.5\%$ to $80.3\%$.

\section{Additional Ablation Studies}
\label{sec:ablation}

\textbf{Additional ablation study.}
We present additional ablation experiments in \cref{tab:additional_ablation} to analyze the effect of backbone training and the loss weighting between depth and canonical map supervision.

First, we evaluate the effect of freezing the backbone. When the backbone is fixed and initialized with the original DA3~\cite{lin2025da3} weights, the model performs poorly on both depth and correspondence estimation, indicating that the pretrained DA3 representation is not sufficient for our task. Using a stronger initialization with a DAViD-pretrained~\cite{saleh2025david} backbone improves performance across all metrics, but still remains noticeably worse than training the backbone jointly with our objectives.

Next, we analyze the impact of the canonical map loss weight $\lambda_C$. When $\lambda_C=1$, the model struggles to learn accurate correspondences because the canonical prediction head must learn the task from scratch, while the depth head benefits from pretrained initialization. Increasing the weight to $\lambda_C=10$ significantly improves correspondence accuracy, achieving the best EPE, but slightly degrades depth performance. Based on this trade-off, we select $\lambda_C=5$ as a balanced setting for our final model.

We further study the effect of disabling the depth supervision by setting $\lambda_D=0$. In this case, the model learns slightly better correspondences but the depth prediction degrades, highlighting the importance of joint geometry supervision. Conversely, removing the canonical loss ($\lambda_C=0$) prevents the model from learning meaningful correspondences, leading to extremely large EPE values, while slightly improving the depth metrics.

Overall, our final model achieves the best balance across the evaluated metrics, obtaining the best average rank while remaining close to the top performance for each individual metric. This demonstrates that our design effectively balances geometry reconstruction and dense correspondence estimation.

\begin{table}[t]
\caption{
\textbf{Additional ablation study.} Component analysis using depth (AbsRel) and correspondence (EPE) evaluation metrics. Best results are shown in \textbf{bold}, second-best \underline{underlined}. AbsRel values are $\times 10$. We also report the average rank across all metrics (lower is better) to summarize overall performance across geometry and correspondence prediction. Our final model achieves the best average rank, indicating the most balanced performance across metrics.
}
\centering
\small
\setlength{\tabcolsep}{7pt}
\renewcommand{\arraystretch}{1.15}
\scalebox{0.84}{
\begin{tabular}{lcccc}
\toprule

\textbf{Method}
& AbsRel (Monocular)$\downarrow$
& EPE$\downarrow$
& AbsRel (16 views)$\downarrow$
& Avg. Rank$\downarrow$ \\

\midrule

Backbone Fixed (DA3)
& 0.101
& 6.9261
& 0.062
& 6.7 \\

\makecell[l]{Backbone Fixed \\ (DAViD Pretrained)}
& 0.082
& 5.8219
& 0.040
& 5.3 \\

$\lambda_C=1$
& \textbf{0.052}
& 4.0316
& \underline{0.015}
& \underline{2.8} \\

$\lambda_C=10$
& 0.056
& \textbf{3.1838}
& \underline{0.015}
& \textbf{2.7} \\

$\lambda_D=0$
& 0.064
& \underline{3.2102}
& 0.060
& 4.3 \\

$\lambda_C=0$
& \textbf{0.052}
& 135.1671
& \textbf{0.014}
& 3.2 \\

Ours
& \underline{0.053}
& 3.271
& \underline{0.015}
& \textbf{2.7} \\

\bottomrule
\end{tabular}
}
\label{tab:additional_ablation}
\end{table}

\textbf{Neighborhood consistency ablation.}
We also evaluate whether adding an explicit neighborhood-consistency loss improves the spatial smoothness of canonical predictions. As shown in \cref{tab:neighborhood_consistency_ablation}, this additional constraint does not improve tracking accuracy on NeRSemble~\cite{kirschstein2023nersemble}. We find that the spatial inductive bias of the DPT prediction head, together with shared transformer features, already encourages locally coherent canonical maps. Adding an explicit neighborhood-consistency term instead slightly over-constrains the predictions, leading to worse metrics.

\begin{table}[t]
\caption{
\textbf{Ablation of neighborhood-consistency loss.}
We evaluate the effect of adding an explicit neighborhood-consistency term for canonical predictions on NeRSemble~\cite{kirschstein2023nersemble}. Best results are shown in \textbf{bold}. The additional loss slightly degrades tracking performance, indicating that the model already learns sufficiently smooth canonical maps without this constraint.
}
\centering
\setlength{\tabcolsep}{3pt}
\renewcommand{\arraystretch}{1.1}
\resizebox{0.98\textwidth}{!}{
\begin{tabular}{lcccc}
\toprule
Method 
& EPE$\downarrow$ 
& $<3$px$\uparrow$ 
& $<5$px$\uparrow$ 
& $<10$px$\uparrow$ \\
\midrule
w/ neighborhood consistency
& 3.43 & 0.58 & 0.81 & \textbf{0.97} \\
Ours
& \textbf{3.30} 
& \textbf{0.60} 
& \textbf{0.83} 
& \textbf{0.97} \\
\bottomrule
\end{tabular}
}
\label{tab:neighborhood_consistency_ablation}
\end{table}

\section{Additional Results}
\label{sec:additional_results}

We first present a comparison to P3DMM~\cite{giebenhain2025pixel3dmm} on CelebV-Text~\cite{yu2023celebv} to illustrate the representational differences between FLAME-based UV predictions and our canonical 3D coordinate representation. We then present additional qualitative results demonstrating the behavior of our method on challenging in-the-wild video frames from VFHQ~\cite{xie2022vfhq}.

\begin{figure}[H]
\centering

\hspace{-17mm} Input Image \hspace{21mm}  P3DMM \hspace{34mm} Ours 

\includegraphics[width=0.98\textwidth]{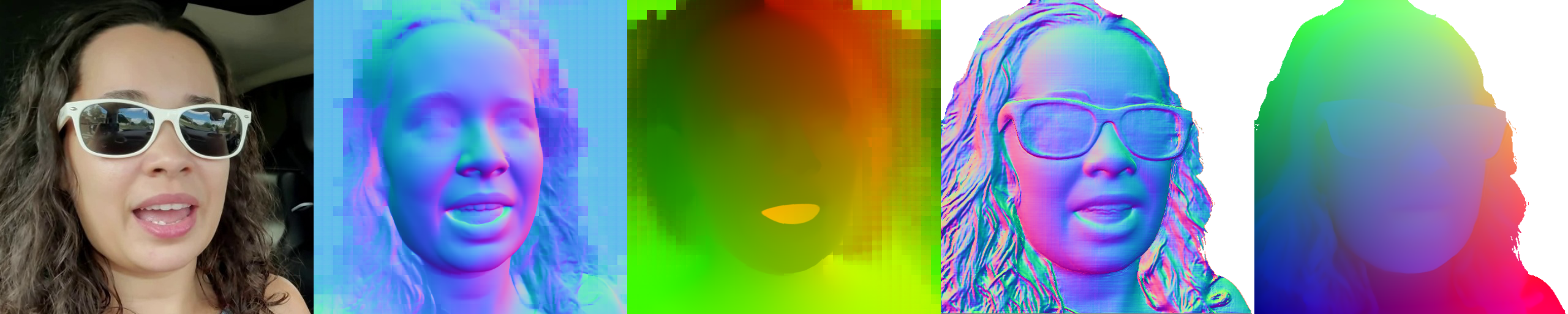}\\[-1mm]

\caption{
\textbf{Normal and canonical map comparison on CelebV-Text.}
We compare P3DMM and our method on an in-the-wild frame from CelebV-Text~\cite{yu2023celebv}. P3DMM predicts normals and 2D UV coordinates constrained by the FLAME surface topology, while our method predicts depth-derived normals and canonical 3D coordinates. Our canonical representation is not restricted to the FLAME mesh and can therefore represent visible regions outside the face topology, such as hair, accessories, and the mouth interior.
}
\label{fig:pixel3dmm_comp}

\end{figure}

\textbf{Novelty with respect to DA3 and P3DMM.}
Our key contribution beyond DA3~\cite{lin2025da3} is reformulating dense 4D face tracking as a canonical-space reconstruction problem, jointly predicting depth, rays, and persistent canonical coordinates, which enables tracking via nearest-neighbor matching. Unlike P3DMM~\cite{giebenhain2025pixel3dmm}, which predicts 2D UV coordinates tied to the FLAME~\cite{li2017flame} surface, our canonical points can occupy arbitrary 3D locations. This allows us to represent regions outside the FLAME topology, such as hair, accessories, and the mouth interior, as shown in \cref{fig:pixel3dmm_comp}. For our method, the shown normals are derived from the predicted depth maps.

\begin{figure}[t]
\centering

\begin{tabular}{@{}cccccc@{}}
\makebox[0.15\textwidth][c]{RGB 1} &
\makebox[0.15\textwidth][c]{Depth 1} &
\makebox[0.15\textwidth][c]{Canonical 1} &
\makebox[0.15\textwidth][c]{RGB 2} &
\makebox[0.15\textwidth][c]{Depth 2} &
\makebox[0.15\textwidth][c]{Canonical 2} \\
\end{tabular}

\includegraphics[width=0.95\textwidth]{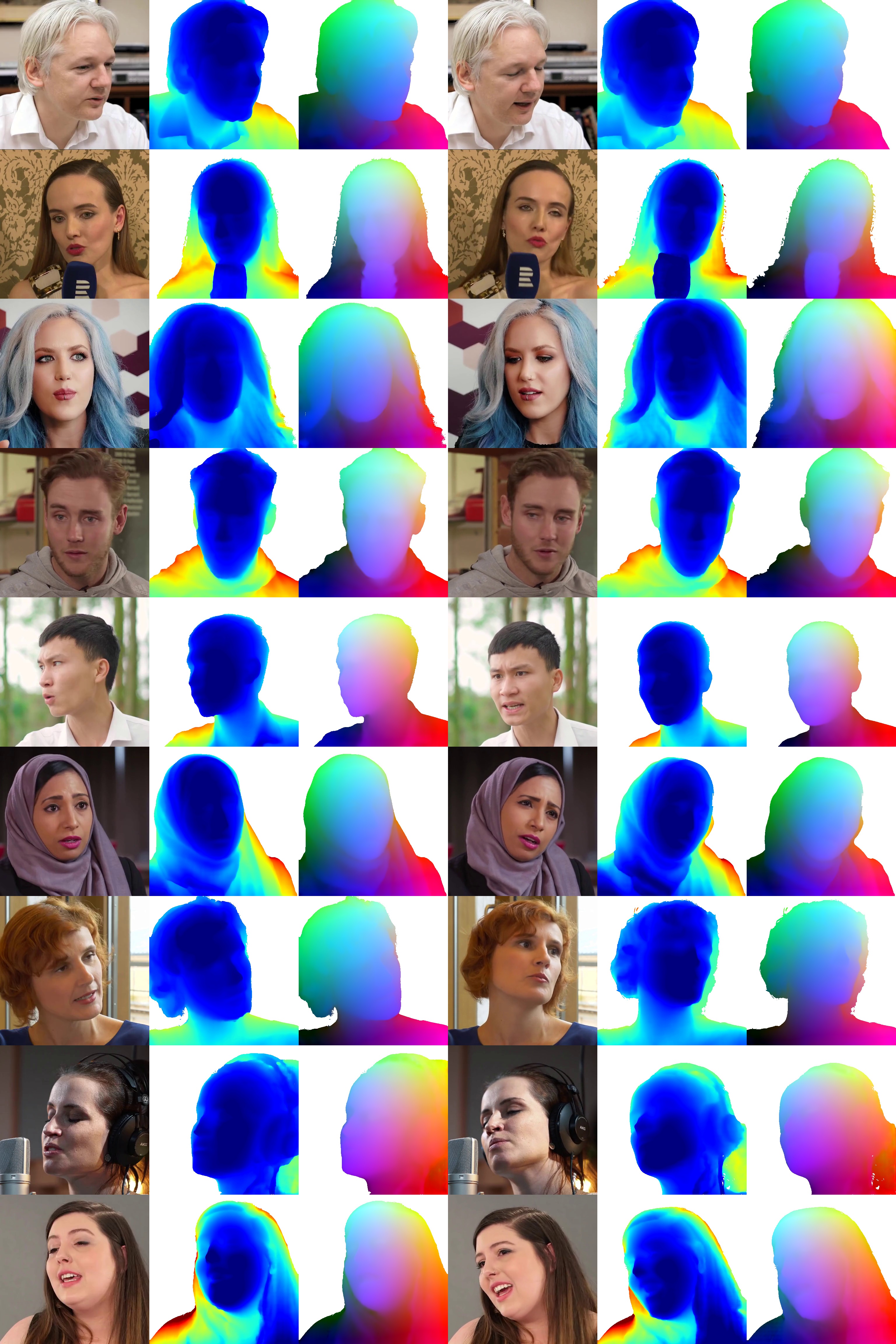}

\caption{
\textbf{Additional prediction examples on VFHQ.}
Given two input views, our method predicts depth maps and canonical maps for each frame.
The results demonstrate consistent geometry and canonical representations across different identities and expressions.
}
\label{fig:additional_examples_vfhq}

\end{figure}

\textbf{Additional depth and canonical predictions.}
\cref{fig:additional_examples_vfhq} shows additional predictions of depth maps and canonical maps from two input views. Despite variations in pose, expression, and identity, the predicted depth and canonical coordinates remain consistent across frames. These results further illustrate the robustness of our approach when reconstructing facial geometry from unconstrained image sequences.

\begin{figure}[t]
\centering

\begin{tabular}{@{}cccccc@{}}
\makebox[0.15\textwidth][c]{RGB 1} &
\makebox[0.30\textwidth][c]{Canonical Point Cloud 1} &
\makebox[0.15\textwidth][c]{RGB 2} &
\makebox[0.30\textwidth][c]{Canonical Point Cloud 2} & \\
\end{tabular}

\includegraphics[width=0.95\textwidth]{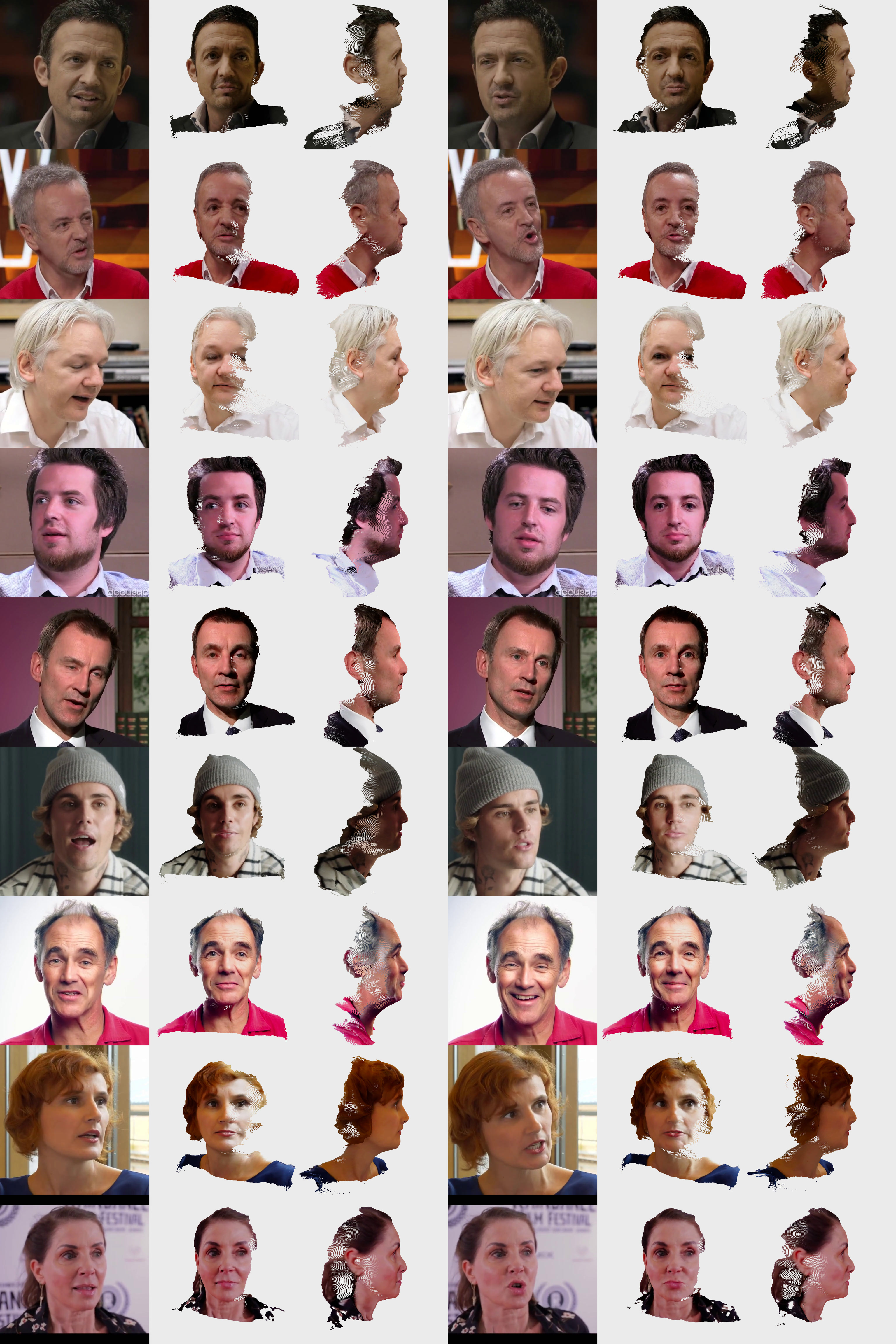}

\caption{
\textbf{Additional canonical point cloud prediction examples on VFHQ.}
Given two input views, our method predicts canonical maps that are backprojected into 3D space to form canonical point clouds. We visualize the reconstructed point clouds from two viewpoints. The results show that the predicted canonical point clouds remain consistent across viewpoints, facial expressions, and identities. The visualization cameras are fixed across all samples.
}
\label{fig:additional_examples_canonical}

\end{figure}

\textbf{Canonical point cloud consistency.}
In \cref{fig:additional_examples_canonical}, we visualize canonical point clouds obtained by backprojecting the predicted canonical maps into 3D space. The resulting point clouds are rendered from two viewpoints while keeping the visualization camera fixed across all samples. The results demonstrate that the canonical representation is consistent across viewpoints, facial expressions, and identities.

\begin{figure}[t]
\centering

\begin{tabular}{@{}cccccc@{}}
\makebox[0.25\textwidth][c]{RGB Images} &
\makebox[0.25\textwidth][c]{4D Reconstruction} &
\makebox[0.25\textwidth][c]{Correspondences} &
\makebox[0.25\textwidth][c]{Canonical Points} & \\
\end{tabular}

\includegraphics[width=1.00\textwidth]{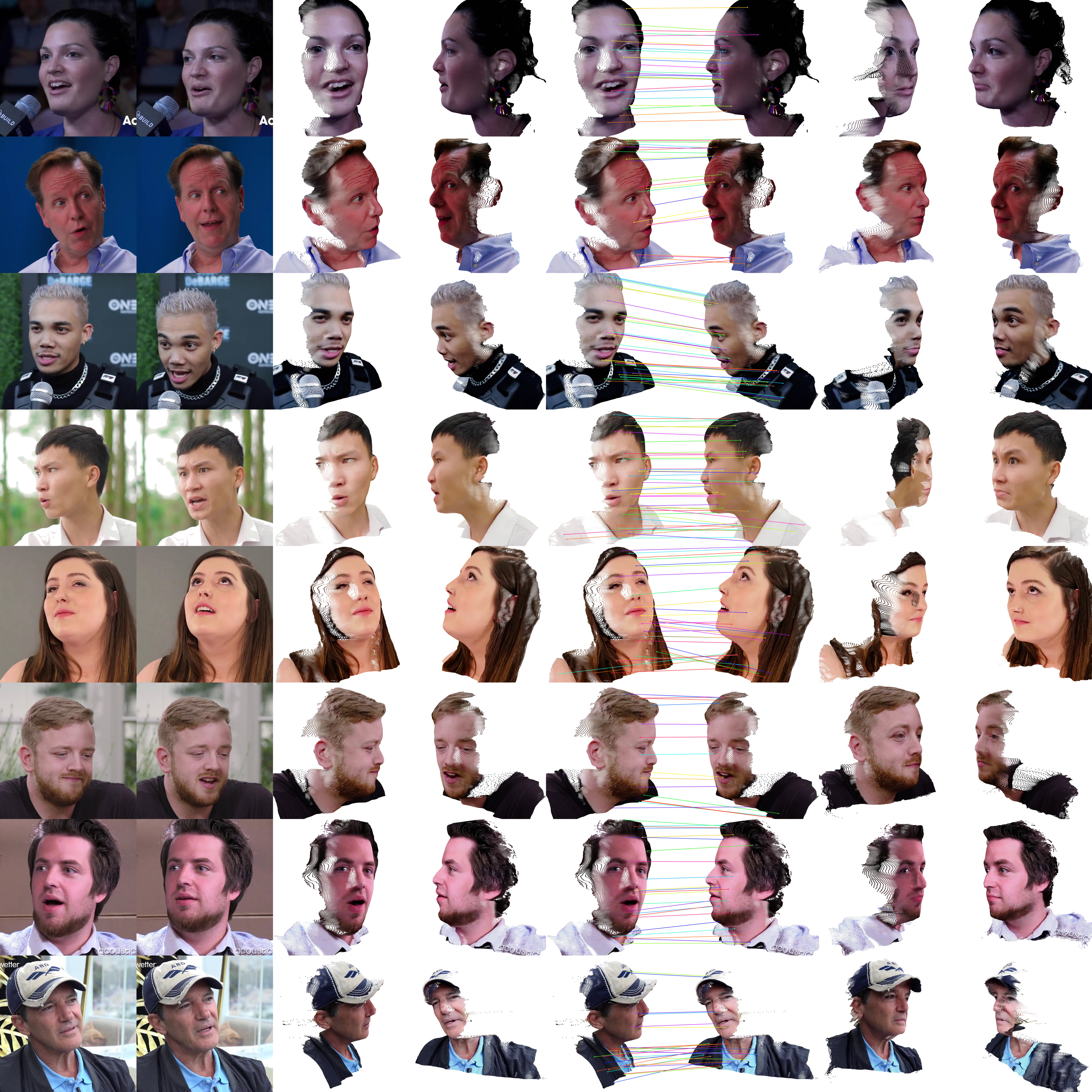}

\caption{
\textbf{Additional predictions on VFHQ.}
Given two RGB input views, our method reconstructs 4D facial geometry and predicts dense correspondences via canonical facial coordinates. The results demonstrate consistent geometry and correspondences across different viewpoints, facial expressions, and identities.
}
\label{fig:additional_examples_teaser}

\end{figure}

\textbf{Additional full pipeline results.}
\cref{fig:additional_examples_teaser} presents further examples of our full pipeline, including the input images, reconstructed geometry, dense correspondences, and canonical point clouds. The predicted correspondences align well across views, highlighting the ability of our method to establish dense facial correspondences while simultaneously reconstructing temporally consistent geometry.

\begin{figure}[t]
\centering

Input Images \hspace{10mm} Predictions \hspace{10mm}  Input Images \hspace{10mm} Predictions 

\includegraphics[width=0.98\textwidth]{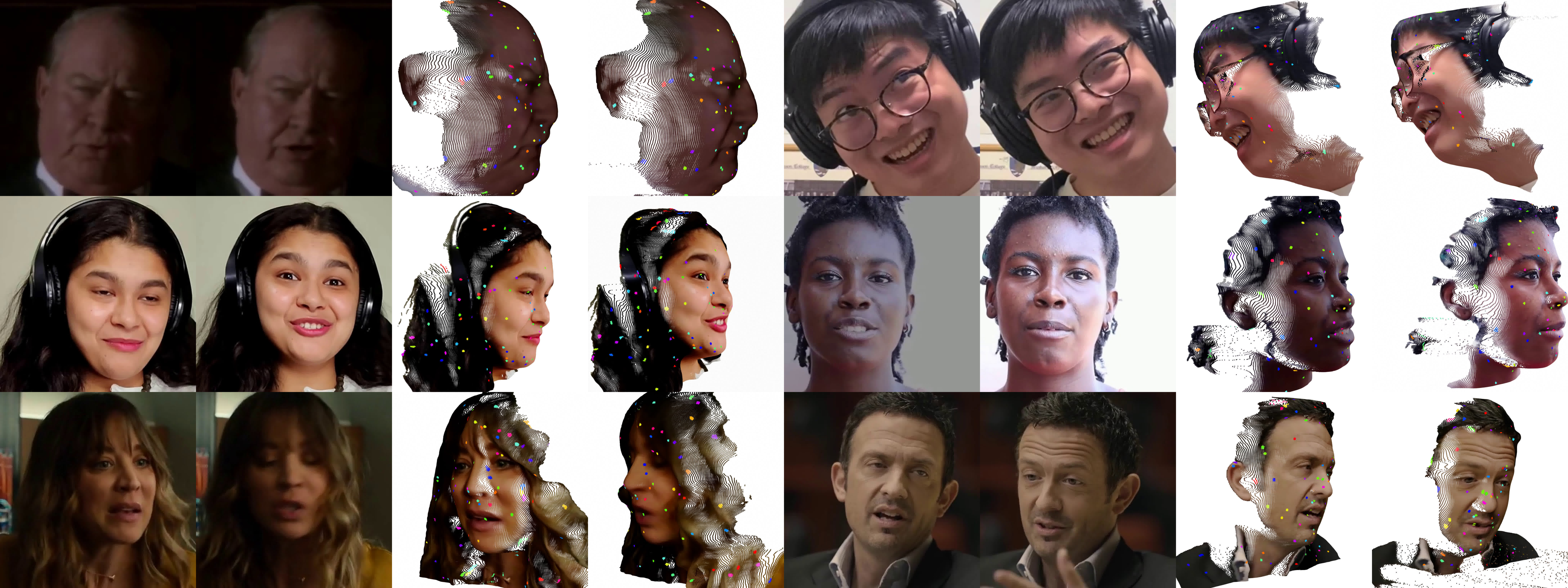}\\[-1mm]

\caption{
\textbf{In-the-wild results on CelebV-HQ~\cite{zhu2022celebvhq}, CelebV-Text~\cite{yu2023celebv}, and Hallo3~\cite{cui2025hallo3}.}
Our method produces stable geometry and correspondence predictions across challenging in-the-wild conditions, including extreme lighting and out-of-distribution identities. The visualized tracks also cover the mouth interior, although fine-grained tracking of individual teeth and tongue motion remains challenging.
}
\label{fig:in_the_wild}

\end{figure}

\textbf{In-the-wild generalization.}
To address the domain gap from NeRSemble~\cite{kirschstein2023nersemble} captures, \cref{fig:in_the_wild} shows in-the-wild results on CelebV-HQ~\cite{zhu2022celebvhq}, CelebV-Text~\cite{yu2023celebv}, and Hallo3~\cite{cui2025hallo3} across challenging scenarios, including extreme lighting and out-of-distribution identities, with colorful tracks including the mouth interior, although per-teeth and tongue tracking remains unresolved. Our model produces stable predictions despite training on studio data, which we attribute to the DA3~\cite{lin2025da3} prior and DAViD~\cite{saleh2025david} pretraining that expose the backbone to diverse in-the-wild imagery and enable our canonical head to generalize beyond NeRSemble.

\begin{table}[t]
\caption{
\textbf{Comparison with 3DGS-based head avatar baselines.}
We compare 2D tracking accuracy on NeRSemble~\cite{kirschstein2023nersemble} against TensorialGaussianAvatar~\cite{wang20253d} and GaussianAvatars~\cite{qian2024gaussianavatars}. Baseline tracks are derived from optimized Gaussian deformations and require per-sequence optimization, while our method performs tracking without test-time optimization. Best results are shown in \textbf{bold}.
}
\centering
\setlength{\tabcolsep}{3pt}
\renewcommand{\arraystretch}{1.1}
\resizebox{0.98\textwidth}{!}{
\begin{tabular}{lcccccc}
\toprule
Method 
& Opt. 
& Time 
& EPE$\downarrow$ 
& $<3$px$\uparrow$ 
& $<5$px$\uparrow$ 
& $<10$px$\uparrow$ \\
\midrule

TensorialGaussianAvatar~\cite{wang20253d}
& Req. 
& $\sim$12h 
& 7.21 & 0.45 & 0.61 & 0.81 \\

GaussianAvatars~\cite{qian2024gaussianavatars}
& Req. 
& $\sim$5h 
& 7.01 & 0.46 & 0.62 & 0.82 \\

Ours
& None 
& 27sec 
& \textbf{5.20} 
& \textbf{0.50} 
& \textbf{0.67} 
& \textbf{0.87} \\

\bottomrule
\end{tabular}
}
\label{tab:gaussian_avatar_comparison}
\end{table}

\textbf{Comparison with 3DGS head avatars.}
We compare 2D tracking accuracy against 3DGS-based dynamic head avatar methods~\cite{qian2024gaussianavatars,wang20253d} on NeRSemble~\cite{kirschstein2023nersemble} in \cref{tab:gaussian_avatar_comparison}. For the baselines, tracks are derived from the optimized Gaussian deformations. Our method outperforms these optimization-based baselines while requiring no per-sequence optimization, demonstrating the robustness and efficiency of canonical-space tracking compared to deformation-based tracking.

\begin{figure}[t]
\centering

\begin{tabular}{@{}cccccc@{}}
\makebox[0.5\textwidth][c]{RGB Images} &
\makebox[0.5\textwidth][c]{Correspondences} &\\
\end{tabular}

\includegraphics[width=1.00\textwidth]{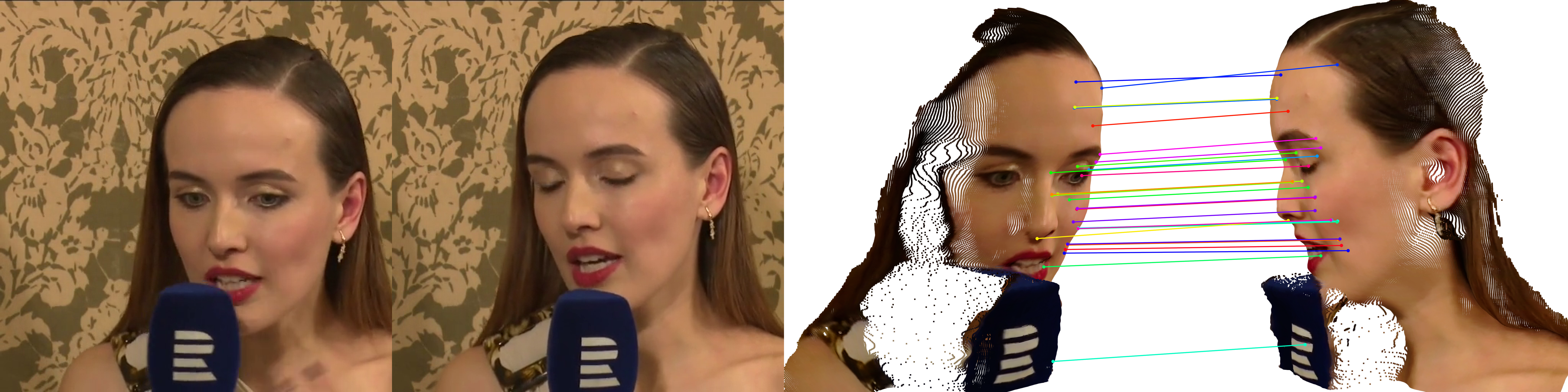}

\caption{
\textbf{Failure case on VFHQ.}
Given two input RGB images, we visualize the predicted correspondences between the reconstructed point clouds. While the correspondences are largely accurate on the facial region, the method fails on the microphone, which is not part of the facial surface and leads to incorrect matches. This highlights a limitation when non-face objects are present in the scene.
}
\label{fig:additional_examples_failure}

\end{figure}

\begin{figure}[t]
\centering

Input Images \hspace{10mm} Predictions \hspace{10mm}  Input Images \hspace{10mm} Predictions 

\includegraphics[width=0.98\textwidth]{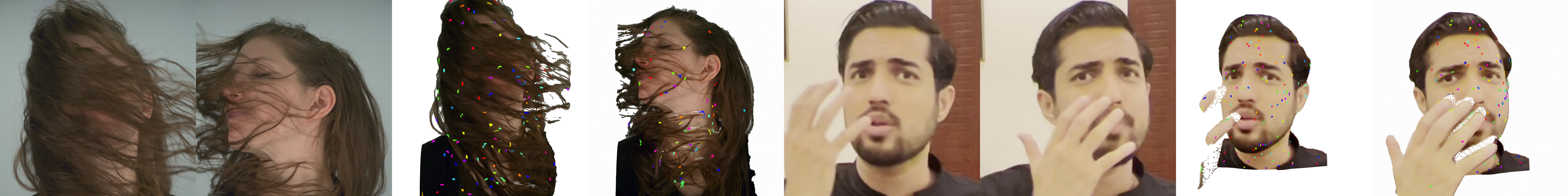}\\[-1mm]

\caption{
\textbf{Failure cases on NeRSemble and CelebV-Text.}
We show representative failure modes on NeRSemble~\cite{kirschstein2023nersemble} and CelebV-Text~\cite{yu2023celebv}. Under extreme hair motion and heavy occlusion, the predicted canonical maps can become unstable, leading to inconsistent geometry and correspondence predictions.
}
\label{fig:failure_cases}

\end{figure}

\textbf{Failure cases.}
We highlight representative failure cases in \cref{fig:additional_examples_failure,fig:failure_cases}. In \cref{fig:additional_examples_failure}, the predicted correspondences remain accurate on the facial region, but the method incorrectly matches the microphone visible in the scene. Since the microphone is not part of the facial surface, this leads to erroneous correspondences when non-face objects appear close to the face. In \cref{fig:failure_cases}, we show additional failure modes under extreme hair motion and heavy occlusion, where the canonical predictions become unstable.

%% file: main.bib
@String(IJCV  = {Int. J. Comput. Vis.})

@String(CVPR  = {IEEE Conf. Comput. Vis. Pattern Recog.})

@String(ICCV  = {Int. Conf. Comput. Vis.})

@String(ECCV  = {Eur. Conf. Comput. Vis.})

@String(NeurIPS = {Adv. Neural Inform. Process. Syst.})

@String(ICLR  = {Int. Conf. Learn. Represent.})

@String(CVPRW = {IEEE Conf. Comput. Vis. Pattern Recog. Worksh.})

@String(AAAI  = {AAAI})

@String(IJCV  = {IJCV})

@String(CVPR  = {CVPR})

@String(ICCV  = {ICCV})

@String(ECCV  = {ECCV})

@String(NeurIPS = {NeurIPS})

@String(ICLR  = {ICLR})

@String(CVPRW = {CVPRW})

@inproceedings{schoenberger2016sfm,
    author={Sch\"{o}nberger, Johannes Lutz and Frahm, Jan-Michael},
    title={{Structure-from-Motion Revisited}},
    booktitle={CVPR},
    year={2016},
}

@inproceedings{schoenberger2016mvs,
    author={Sch\"{o}nberger, Johannes Lutz and Zheng, Enliang and Pollefeys, Marc and Frahm, Jan-Michael},
    title={{Pixelwise View Selection for Unstructured Multi-View Stereo}},
    booktitle={ECCV},
    year={2016},
}

@inproceedings{mildenhall2020nerf,
 title={{NeRF: Representing Scenes as Neural Radiance Fields for View Synthesis}},
 author={Ben Mildenhall and Pratul P. Srinivasan and Matthew Tancik and Jonathan T. Barron and Ravi Ramamoorthi and others},
 year={2020},
 booktitle={ECCV},
}

@inproceedings{martinbrualla2020nerfw,
title = {{NeRF in the Wild: Neural Radiance Fields for Unconstrained Photo Collections}},
author = {Martin-Brualla, Ricardo and Radwan, Noha and Sajjadi, Mehdi S. M. and Barron, Jonathan T. and Dosovitskiy, Alexey and others},
booktitle = {CVPR},
year={2021}
}

@inproceedings{wang2022sparsenerf,
   author    = {Wang, Guangcong and Chen, Zhaoxi and Loy, Chen Change and Liu, Ziwei},
   title     = {{SparseNeRF: Distilling Depth Ranking for Few-shot Novel View Synthesis}},
   booktitle = {ICCV},   
   year      = {2023},
  }

@inproceedings{wang2023dust3r,
  title={{DUSt3R: Geometric 3D Vision Made Easy}},
  author={Shuzhe Wang and Vincent Leroy and Yohann Cabon and Boris Chidlovskii and J{\'e}r{\^o}me Revaud},
  booktitle={CVPR},
  year={2024}, 
}

@inproceedings{leroy2024mast3r,
      title={{Grounding Image Matching in 3D with MASt3R}}, 
      author={Vincent Leroy and Yohann Cabon and Jerome Revaud},
      booktitle = {ECCV},
      year = {2024}
}

@inproceedings{charatan2023pixelsplat,
      title={{pixelSplat: 3D Gaussian Splats from Image Pairs for Scalable Generalizable 3D Reconstruction}},
      author={David Charatan and Sizhe Li and Andrea Tagliasacchi and Vincent Sitzmann},
      year={2024},
      booktitle={CVPR},
}

@inproceedings{xu2024depthsplat,
      title   = {{DepthSplat: Connecting Gaussian Splatting and Depth}},
      author  = {Xu, Haofei and Peng, Songyou and Wang, Fangjinhua and Blum, Hermann and Barath, Daniel and others},
      booktitle={CVPR},
      year={2025}
    }

@inproceedings{chen2024mvsplat,
    title   = {{MVSplat: Efficient 3D Gaussian Splatting from Sparse Multi-View Images}},
    author  = {Chen, Yuedong and Xu, Haofei and Zheng, Chuanxia and Zhuang, Bohan and Pollefeys, Marc and others},
    booktitle = {ECCV},
    year    = {2024},
}

@inproceedings{wang2025pi,
  title={{Pi3: Permutation-Equivariant Visual Geometry Learning}},
  author={Wang, Yifan and Zhou, Jianjun and Zhu, Haoyi and Chang, Wenzheng and Zhou, Yang and others},
  booktitle={ICLR},
  year={2026}
}

@inproceedings{wang2025vggt,
  title={{VGGT: Visual Geometry Grounded Transformer}},
  author={Wang, Jianyuan and Chen, Minghao and Karaev, Nikita and Vedaldi, Andrea and Rupprecht, Christian and others},
  booktitle={CVPR},
  year={2025}
}

@inproceedings{pumarola2020dnerf,
    title={{D-NeRF: Neural Radiance Fields for Dynamic Scenes}},
    author={Pumarola, Albert and Corona, Enric and Pons-Moll, Gerard and Moreno-Noguer, Francesc},
    booktitle={CVPR},
    year={2020}
}

@inproceedings{park2021nerfies,
  author    = {Park, Keunhong and Sinha, Utkarsh and Barron, Jonathan T. and Bouaziz, Sofien and Goldman, Dan B and others},
  title     = {{Nerfies: Deformable Neural Radiance Fields}},
  booktitle   = {ICCV},
  year      = {2021},
}

@article{park2021hypernerf,
author = {Park, Keunhong and Sinha, Utkarsh and Hedman, Peter and Barron, Jonathan T. and Bouaziz, Sofien and others},
title = {{HyperNeRF: A Higher-Dimensional Representation for Topologically Varying Neural Radiance Fields}},
journal = {ACM Trans. Graph.},
issue_date = {December 2021},
publisher = {ACM},
volume = {40},
number = {6},
month = {dec},
year = {2021},
articleno = {238},
}

@article{sucar2025vdpm,
  title={{{V-DPM}: 4D Video Reconstruction with Dynamic Point Maps}},
  author={Sucar, Edgar and Insafutdinov, Eldar and Lai, Zihang and Vedaldi, Andrea},
  journal={arXiv preprint arXiv:2601.09499},
  year={2025}
}

@inproceedings{
bian2023contextpips,
title={{Context-PIPs: Persistent Independent Particles Demands Context Features}},
author={Weikang Bian and Zhaoyang Huang and Xiaoyu Shi and Yitong Dong and Yijin Li and others},
booktitle={NeurIPS},
year={2023},
}

@inproceedings{doersch2023tapir,
  title={{{TAPIR}: Tracking Any Point with per-frame Initialization and Temporal Refinement}},
  author={Doersch, Carl and Yang, Yi and Vecerik, Mel and Gokay, Dilara and Gupta, Ankush and others},
  booktitle={ICCV},
  year={2023}
}

@inproceedings{karaev23cotracker,
  title     = {{CoTracker: It is Better to Track Together}},
  author    = {Nikita Karaev and Ignacio Rocco and Benjamin Graham and Natalia Neverova and Andrea Vedaldi and others},
  booktitle = {ECCV},
  year      = {2024}
}

@inproceedings{li2024taptr,
  title={{TAPTR: Tracking Any Point with Transformers as Detection}},
  author={Li, Hongyang and Zhang, Hao and Liu, Shilong and Zeng, Zhaoyang and Ren, Tianhe and others},
  booktitle={ECCV},
  year={2024},
}

@inproceedings{cho2024locotrack,
    title={{Local All-Pair Correspondence for Point Tracking}}, 
    author={Seokju Cho and Jiahui Huang and Jisu Nam and Honggyu An and Seungryong Kim and others},
    booktitle={ECCV},
    year={2024},
}

@article{li2017flame, 
  title = {{Learning a Model of Facial Shape and Expression from {4D} Scans}}, 
  author = {Li, Tianye and Bolkart, Timo and Black, Michael. J. and Li, Hao and Romero, Javier}, 
  journal = {ACM Trans. Graph., (Proc. SIGGRAPH Asia)}, 
  volume = {36}, 
  number = {6}, 
  year = {2017}, 
  pages = {194:1--194:17},
}

@inproceedings{danecek2022emoca,
  title = {{{EMOCA}: {E}motion Driven Monocular Face Capture and Animation}},
  author = {Danecek, Radek and Black, Michael J. and Bolkart, Timo},
  booktitle = {CVPR},
  year = {2022}
}

@article{giebenhain2025pixel3dmm,
title={{Pixel3DMM: Versatile Screen-Space Priors for Single-Image 3D Face Reconstruction}},
author={Giebenhain, Simon and Kirschstein, Tobias and R{\"u}nz, Tobias and Agapito, Lourdes and Nie{\ss}ner, Matthias},
journal={arXiv preprint arXiv:2505.00615},
year={2025}
}

@InProceedings{wang2022faceverse,
title={{FaceVerse: a Fine-grained and Detail-controllable 3D Face Morphable Model from a Hybrid Dataset}},
author={Wang, Lizhen and Chen, Zhiyua and Yu, Tao and Ma, Chenguang and Li, Liang and Liu, Yebin},
booktitle={CVPR},
month={June},
year={2022},
}

@inproceedings{zheng2022imavatar,
  title={{IM Avatar: Implicit Morphable Head Avatars from Videos}},
  author={Zheng, Yufeng and Abrevaya, Victoria Fern{\'a}ndez and B{\"u}hler, Marcel C. and Chen, Xu and Black, Michael J. and Hilliges, Otmar},
  booktitle = {CVPR},
  year = {2022}
}

@inproceedings{grassal2022neural,
title={{Neural Head Avatars from Monocular RGB Videos}},
author={Grassal, Philip-William and Prinzler, Malte and Leistner, Titus and Rother, Carsten and Nie{\ss}ner, Matthias and others},
booktitle={CVPR},
year={2022}
}

@inproceedings{paysan2009bfm,
title={{A 3D Face Model for Pose and Illumination Invariant Face Recognition}},
author={P. Paysan and R. Knothe and B. Amberg and S. Romdhani and T. Vetter},
booktitle={IEEE International Conference on Advanced Video and Signal based Surveillance (AVSS) for Security, Safety and Monitoring in Smart Environments},
year={2009},
}

@inproceedings{gerig2018face,
title={{Morphable Face Models - An Open Framework}},
author={Thomas Gerig and Andreas Morel-Forster and Clemens Blumer and Bernhard Egger and Marcel L{\"u}thi and others},
booktitle={IEEE Conference on Automatic Face and Gesture Recognition},
pages={75--82},
year={2018}
}

@inproceedings{lihe2024da,
      title={{Depth Anything: Unleashing the Power of Large-Scale Unlabeled Data}}, 
      author={Yang, Lihe and Kang, Bingyi and Huang, Zilong and Xu, Xiaogang and Feng, Jiashi and Zhao, Hengshuang},
      booktitle={CVPR},
      year={2024}
}

@article{lihe2024da2,
  title={{Depth Anything V2}},
  author={Yang, Lihe and Kang, Bingyi and Huang, Zilong and Zhao, Zhen and Xu, Xiaogang and Feng, Jiashi and Zhao, Hengshuang},
  journal={arXiv preprint arXiv:2406.09414},
  year={2024}
}

@article{lin2025da3,
  title={{Depth Anything 3: Recovering the Visual Space from Any Views}},
  author={Haotong Lin and Sili Chen and Jun Hao Liew and Donny Y. Chen and Zhenyu Li and Guang Shi and Jiashi Feng and Bingyi Kang},
  journal={arXiv preprint arXiv:2511.10647},
  year={2025}
}

@inproceedings{moreau2026offthegrid,
      title={{Off The Grid: Detection of Primitives for Feed-Forward 3D Gaussian Splatting}},
      author={Moreau, Arthur and Shaw, Richard and Nazarczuk, Michal  and Shin, Jisu and Tanay, Thomas and others},
      booktitle={CVPR},
      year={2026}
    }

@Article{kerbl3Dgaussians,
      author       = {Kerbl, Bernhard and Kopanas, Georgios and Leimk{\"u}hler, Thomas and Drettakis, George},
      title        = {{3D Gaussian Splatting for Real-Time Radiance Field Rendering}},
      journal      = {ACM Trans. Graph.},
      number       = {4},
      volume       = {42},
      month        = {July},
      year         = {2023},
}

@inproceedings{chen2022tensorf,
              author = {Anpei Chen and Zexiang Xu and Andreas Geiger and Jingyi Yu and Hao Su},
              title = {{TensoRF: Tensorial Radiance Fields}},
              booktitle = {ECCV},
              year = {2022}
            }

@inproceedings{barron2021mipnerf,
    title={{Mip-NeRF: A Multiscale Representation 
           for Anti-Aliasing Neural Radiance Fields}},
    author={Jonathan T. Barron and Ben Mildenhall and 
            Matthew Tancik and Peter Hedman and 
            Ricardo Martin-Brualla and others},
    booktitle={ICCV},
    year={2021}
}

@article{mueller2022instant,
    author = {Thomas M\"uller and Alex Evans and Christoph Schied and Alexander Keller},
    title = {{Instant Neural Graphics Primitives with a Multiresolution Hash Encoding}},
    journal = {ACM Trans. Graph.},
    issue_date = {July 2022},
    volume = {41},
    number = {4},
    month = jul,
    year = {2022},
    pages = {102:1--102:15},
    articleno = {102},
    numpages = {15},
    publisher = {ACM},
    address = {New York, NY, USA}
}

@inproceedings{niemeyer2021regnerf,
          author    = {Michael Niemeyer and Jonathan T. Barron and Ben Mildenhall and Mehdi S. M. Sajjadi and Andreas Geiger and others},  
          title     = {{RegNeRF: Regularizing Neural Radiance Fields for View Synthesis from Sparse Inputs}},
          booktitle = {CVPR},
          year      = {2022},
        }

@inproceedings{chen2025dashgaussian,
  title     = {{DashGaussian: Optimizing 3D Gaussian Splatting in 200 Seconds}},
  author    = {Chen, Youyu and Jiang, Junjun and Jiang, Kui and Tang, Xiao and Li, Zhihao and others},
  booktitle = {CVPR},
  year      = {2025}
}

@inproceedings{mallick2024taming3dgs,
    author = {Mallick, Saswat Subhajyoti and Goel, Rahul and Kerbl, Bernhard and Steinberger, Markus and Carrasco, Francisco Vicente and others},
    title = {{Taming 3DGS: High-Quality Radiance Fields with Limited Resources}},
    year = {2024},
    publisher = {Association for Computing Machinery},
    address = {New York, NY, USA},
    booktitle = {SIGGRAPH Asia},
    articleno = {2},
    numpages = {11},
    }

@article{meuleman2025onthefly,
  title={{On-the-fly Reconstruction for Large-Scale Novel View Synthesis from Unposed Images}},
  author={Meuleman, Andreas and Shah, Ishaan and Lanvin, Alexandre and Kerbl, Bernhard and Drettakis, George},
  journal={ACM Trans. Graph.},
  volume={44},
  number={4},
  year={2025}
}

@article{jiang2025anysplat,
  title={{AnySplat: Feed-forward 3d Gaussian Splatting from Unconstrained Views}},
  author={Jiang, Lihan and Mao, Yucheng and Xu, Linning and Lu, Tao and Ren, Kerui and others},
  journal={ACM Trans. Graph.},
  volume={44},
  number={6},
  pages={1--16},
  year={2025},
  publisher={ACM New York, NY, USA}
}

@inproceedings{keetha2026mapanything,
  title={{MapAnything: Universal Feed-Forward Metric {3D} Reconstruction}},
  author={Nikhil Keetha and Norman M\"{u}ller and Johannes Sch\"{o}nberger and Lorenzo Porzi and Yuchen Zhang and others},
  booktitle={3DV},
  year={2026},
}

@inproceedings{cut3r,
author = {Qianqian Wang and Yifei Zhang and Aleksander Holynski and Alexei A. Efros and Angjoo Kanazawa},
title = {{Continuous 3D Perception Model with Persistent State}},
year = {2025},
booktitle={CVPR},
}

@InProceedings{wu20244dgs,
    author    = {Wu, Guanjun and Yi, Taoran and Fang, Jiemin and Xie, Lingxi and Zhang, Xiaopeng and others},
    title     = {{4D Gaussian Splatting for Real-Time Dynamic Scene Rendering}},
    booktitle = {CVPR},
    year      = {2024},
}

@inproceedings{Shaw2024swings,
  title={{SWinGS: Sliding Windows for Dynamic 3D Gaussian Splatting}},
  author={Richard Shaw and Jifei Song and Arthur Moreau and Michal Nazarczuk and Sibi Catley-Chandar and others},
  booktitle={ECCV},
  year={2024},
}

@inproceedings{li2023spacetime,
  title={{Spacetime Gaussian Feature Splatting for Real-Time Dynamic View Synthesis}},
  author={Li, Zhan and Chen, Zhang and Li, Zhong and Xu, Yi},
  booktitle={CVPR},
  year={2024}
}

@inproceedings{zhang2024monst3r,
  title={{MonST3R: A Simple Approach for Estimating Geometry in the Presence of Motion}},
  author={Zhang, Junyi and Herrmann, Charles and Hur, Junhwa and Jampani, Varun and Darrell, Trevor and others},
  booktitle={ICLR},
  year={2025}
}

@inproceedings{lu2024align3r,
  title={{Align3R: Aligned Monocular Depth Estimation for Dynamic Videos}},
  author={Lu, Jiahao and Huang, Tianyu and Li, Peng and Dou, Zhiyang and Lin, Cheng and others},
  booktitle={CVPR},
  year={2025}
}

@inproceedings{zhou2026page4d,
  title={{PAGE-4D: Disentangled Pose and Geometry Estimation for 4D Perception}},
  author={Zhou, Kaichen and Zhou, Kaichen and Wang, Yuhan and Chen, Grace and Beaudouin, Gaspard and others},
  booktitle={ICLR},
  year={2026}
}

@inproceedings{jiang2025geo4d,
      title={{Geo4D: Leveraging Video Generators for Geometric 4D Scene Reconstruction}}, 
      author={Zeren Jiang and Chuanxia Zheng and Iro Laina and Diane Larlus and Andrea Vedaldi},
      booktitle={ICCV},
      year={2025},
}

@inproceedings{sucar2025dynamicpm,
  title={{Dynamic Point Maps: A Versatile Representation for Dynamic 3D Reconstruction}},
  author={Edgar Sucar and Zihang Lai and Eldar Insafutdinov and Andrea Vedaldi},
  booktitle={ICCV},
  year={2025},
}

@inproceedings{st4rtrack2025,
  title={{St4RTrack: Simultaneous 4D Reconstruction and Tracking in the World}},
  author={Feng, Haiwen and Zhang, Junyi and Wang, Qianqian and Ye, Yufei and Yu, Pengcheng and others},
  booktitle={ICCV},
  year={2025}
}

@inproceedings{luiten2023dynamic,
  title={{Dynamic 3D Gaussians: Tracking by Persistent Dynamic View Synthesis}},
  author={Luiten, Jonathon and Kopanas, Georgios and Leibe, Bastian and Ramanan, Deva},
  booktitle={3DV},
  year={2024}
}

@inproceedings{saleh2025david,
  title={{David: Data-efficient and Accurate Vision Models from Synthetic Data}},
  author={Saleh, Fatemeh and Aliakbarian, Sadegh and Hewitt, Charlie and Petikam, Lohit and Xiao, Xian and others},
  booktitle={ICCV},
  year={2025}
}

@inproceedings{khirodkar2024sapiens,
  title={{Sapiens: Foundation for Human Vision Models}},
  author={Khirodkar, Rawal and Bagautdinov, Timur and Martinez, Julieta and Zhaoen, Su and James, Austin and others},
  booktitle={ECCV},
  year={2024},
}

@inproceedings{ranftl2021vision,
  title={{Vision Transformers for Dense Prediction}},
  author={Ranftl, Ren{\'e} and Bochkovskiy, Alexey and Koltun, Vladlen},
  booktitle={ICCV},
  year={2021}
}

@article{kirschstein2023nersemble,
  title={{NeRSemble: Multi-view Radiance Field Reconstruction of Human Heads}},
  author={Kirschstein, Tobias and Qian, Shenhan and Giebenhain, Simon and Walter, Tim and Nie{\ss}ner, Matthias},
  journal={ACM Trans. Graph.},
  volume={42},
  number={4},
  pages={1--14},
  year={2023},
  publisher={ACM New York, NY, USA}
}

@article{lugaresi2019mediapipe,
  title={{MediaPipe: A Framework for Building Perception Pipelines}},
  author={Lugaresi, Camillo and Tang, Jiuqiang and Nash, Hadon and McClanahan, Chris and Uboweja, Esha and others},
  journal={arXiv preprint arXiv:1906.08172},
  year={2019}
}

@inproceedings{huang2023sc,
        title={{SC-GS: Sparse-Controlled Gaussian Splatting for Editable Dynamic Scenes}},
        author={Huang, Yi-Hua and Sun, Yang-Tian and Yang, Ziyi and Lyu, Xiaoyang and Cao, Yan-Pei and others},
        booktitle={CVPR},
        year={2024}
      }

@inproceedings{blanz1999morphable,
  title={{A Morphable Model for the Synthesis of 3D Faces}},
  author={Blanz, Volker and Vetter, Thomas},
  booktitle={SIGGRAPH},
  year={1999}
}

@inproceedings{tewari2017mofa,
  title={{MoFA: Model-Based Deep Convolutional Face Autoencoder for Unsupervised Monocular Reconstruction}},
  author={Ayush Kumar Tewari and Michael Zollh{\"o}fer and Hyeongwoo Kim and Pablo Garrido and Florian Bernard and others},
  booktitle={ICCV},
  year={2017},
}

@inproceedings{richardson20173d,
  title={{Learning Detailed Face Reconstruction from a Single Image}},
  author={Elad Richardson and Matan Sela and Roy Or-El and Ron Kimmel},
  booktitle={CVPR},
  year={2017}
}

@inproceedings{deng2019accurate,
  title={{Accurate 3D Face Reconstruction With Weakly-Supervised Learning: From Single Image to Image Set}},
  author={Yu Deng and Jiaolong Yang and Sicheng Xu and Dong Chen and Yunde Jia and others},
  booktitle={CVPRW},
  year={2019},
}

@article{martinez2024codec,
  author = {Julieta Martinez and Emily Kim and Javier Romero and Timur Bagautdinov and Shunsuke Saito and others},
  title = {{Codec Avatar Studio: Paired Human Captures for Complete, Driveable, and Generalizable Avatars}},
  year = {2024},
  journal = {NeurIPS Track on Datasets and Benchmarks},
}

@InProceedings{xie2022vfhq,
      author = {Liangbin Xie and Xintao Wang and Honglun Zhang and Chao Dong and Ying Shan},
      title = {{VFHQ: A High-Quality Dataset and Benchmark for Video Face Super-Resolution}},
      booktitle={CVPRW},
      year = {2022}
  }

@inproceedings{zhu2022celebvhq,
  title={{{CelebV-HQ}: A Large-Scale Video Facial Attributes Dataset}},
  author={Zhu, Hao and Wu, Wayne and Zhu, Wentao and Jiang, Liming and Tang, Siwei and others},
  booktitle={ECCV},
  year={2022}
}

@inproceedings{dhamo2023headgas,
  title={{HeadGaS: Real-Time Animatable Head Avatars via 3D Gaussian Splatting}},
  author={Dhamo, Helisa and Nie, Yinyu and Moreau, Arthur and Song, Jifei and Shaw, Richard and others},
  booktitle={ECCV},
  year={2024}
}

@article{zhao2024psavatar,
  title={{PSAvatar: A Point-based Shape Model for Real-Time Head Avatar Animation with 3D Gaussian Splatting}},
  author={Zhongyuan Zhao and Zhenyu Bao and Qing Li and Guoping Qiu and Kanglin Liu},
  journal={arXiv preprint arXiv:2401.12900},
  year={2024},
}

@inproceedings{xu2023gaussianheadavatar,
  title={{Gaussian Head Avatar: Ultra High-fidelity Head Avatar via Dynamic Gaussians}},
  author={Xu, Yuelang and Chen, Benwang and Li, Zhe and Zhang, Hongwen and Wang, Lizhen and others},
  booktitle={CVPR},
  year={2024}
}

@article{wang2024gaussianhead,
    title={{GaussianHead: High-fidelity Head Avatars with Learnable Gaussian Derivation}}, 
    author={Jie Wang and Jiu-Cheng Xie and Xianyan Li and Feng Xu and Chi-Man Pun and others},
    journal={IEEE Trans. on Visualization and Computer Graphics},
    year={2025},
}

@article{chen2024monogaussianavatar,
  title={{MonoGaussianAvatar: Monocular Gaussian Point-based Head Avatar}},
  author={Chen, Yufan and Wang, Lizhen and Li, Qijing and Xiao, Hongjiang and Zhang, Shengping and others},
  journal={ACM Trans. Graph., Proc. SIGGRAPH},
  year={2024}
}

@inproceedings{zhou2024headstudio,
  title = {{HeadStudio: Text to Animatable Head Avatars with 3D Gaussian Splatting}},
  author = {Zhenglin Zhou and Fan Ma and Hehe Fan and Zongxin Yang and Yi Yang},
  booktitle = {ECCV},
  year={2024},
}

@article{shaw2025ico3d,
      title={{ICo3D: An Interactive Conversational 3D Virtual Human}},
      author={Shaw, Richard and Jang, Youngkyoon and Papaioannou, Athanasios and Moreau, Arthur and Dhamo, Helisa and others},
      journal={IJCV},
      year={2025}
    }

@inproceedings{xu20254dgt,
    title = {{4DGT: Learning a 4D Gaussian Transformer Using Real-World Monocular Videos}},
    author = {Xu, Zhen and Li, Zhengqin and Dong, Zhao and Zhou, Xiaowei and Newcombe, Richard and others},
    booktitle = {NeurIPS},
    year = {2025}
}

@inproceedings{wang2025movies,
  title={{MoVieS: Motion-Aware 4D Dynamic View Synthesis in One Second}},
  author={Lin, Chenguo and Lin, Yuchen and Pan, Panwang and Yu, Yifan and Hu, Tao and others},
  booktitle={CVPR},
  year={2026}
}

@inproceedings{wu2026streamsplat,
    title={{StreamSplat: Towards Online Dynamic 3D Reconstruction from Uncalibrated Video Streams}}, 
    author={Zike Wu and Qi Yan and Xuanyu Yi and Lele Wang and Renjie Liao},
    booktitle={ICLR},
    year={2026},
}

@inproceedings{ye2024noposplat,
      title   = {{No Pose, No Problem: Surprisingly Simple 3D Gaussian Splats from Sparse Unposed Images}},
      author  = {Ye, Botao and Liu, Sifei and Xu, Haofei and Xueting, Li and Pollefeys, Marc and others},
      booktitle = {ICLR},
      year    = {2025}
    }

@article{bentley1975multidimensional,
      title={{Multidimensional Binary Search Trees used for Associative Searching}},
  author={Bentley, Jon Louis},
  journal={Communications of the ACM},
  volume={18},
  number={9},
  pages={509--517},
  year={1975},
  publisher={ACM}
}

@inproceedings{feng2018prnet,
title={{Joint 3D Face Reconstruction and Dense Alignment with Position Map Regression Network}},
author={Yao Feng and Fan Wu and Xiaohu Shao and Yanfeng Wang and Xi Zhou},
booktitle={ECCV},
year={2018},
}

@inproceedings{wu2026uika,
  title={{UIKA: Fast Universal Head Avatar from Pose-Free Images}},
  author={Zijian Wu and Boyao Zhou and Liangxiao Hu and Hongyu Liu and Yuan Sun and others},
  booktitle={CVPR},
  year={2026},
}

@inproceedings{ming2025vggtface,
  title={{VGGTFace: Topologically Consistent Facial Geometry Reconstruction in the Wild}},
  author={Xin Ming and Yuxuan Han and Tianyu Huang and Feng Xu},
  booktitle={AAAI},
  year={2026},
}

@inproceedings{adamW2019,
  title={{Decoupled Weight Decay Regularization}},
  author={Ilya Loshchilov and Frank Hutter},
  booktitle={ICLR},
  year={2019},
}

@inproceedings{zheng2022farl,
  title={General facial representation learning in a visual-linguistic manner},
  author={Zheng, Yinglin and Yang, Hao and Zhang, Ting and Bao, Jianmin and Chen, Dongdong and Huang, Yangyu and Yuan, Lu and Chen, Dong and Zeng, Ming and Wen, Fang},
  booktitle={Proceedings of the IEEE/CVF Conference on Computer Vision and Pattern Recognition},
  pages={18697--18709},
  year={2022}
}

@inproceedings{yu2023celebv,
  title={Celebv-text: A large-scale facial text-video dataset},
  author={Yu, Jianhui and Zhu, Hao and Jiang, Liming and Loy, Chen Change and Cai, Weidong and Wu, Wayne},
  booktitle={Proceedings of the IEEE/CVF Conference on Computer Vision and Pattern Recognition},
  pages={14805--14814},
  year={2023}
}

@inproceedings{cui2025hallo3,
  title={Hallo3: Highly dynamic and realistic portrait image animation with video diffusion transformer},
  author={Cui, Jiahao and Li, Hui and Zhan, Yun and Shang, Hanlin and Cheng, Kaihui and Ma, Yuqi and Mu, Shan and Zhou, Hang and Wang, Jingdong and Zhu, Siyu},
  booktitle={Proceedings of the Computer Vision and Pattern Recognition Conference},
  pages={21086--21095},
  year={2025}
}

@inproceedings{wang20253d,
  title={3d gaussian head avatars with expressive dynamic appearances by compact tensorial representations},
  author={Wang, Yating and Wang, Xuan and Yi, Ran and Fan, Yanbo and Hu, Jichen and Zhu, Jingcheng and Ma, Lizhuang},
  booktitle={Proceedings of the Computer Vision and Pattern Recognition Conference},
  pages={21117--21126},
  year={2025}
}

@inproceedings{qian2024gaussianavatars,
  title={Gaussianavatars: Photorealistic head avatars with rigged 3d gaussians},
  author={Qian, Shenhan and Kirschstein, Tobias and Schoneveld, Liam and Davoli, Davide and Giebenhain, Simon and Nie{\ss}ner, Matthias},
  booktitle={Proceedings of the IEEE/CVF Conference on Computer Vision and Pattern Recognition},
  pages={20299--20309},
  year={2024}
}
